\documentclass[10pt,twocolumn,letterpaper]{article}

\usepackage{cvpr}              

\usepackage[T1]{fontenc}
\usepackage{amsmath}
\usepackage{amssymb}
\usepackage{mathtools}
\usepackage{graphicx}
\usepackage{tikz,pgfplots,filecontents}
\usepackage{booktabs}
\usepackage{threeparttable}
\usepackage{multirow}
\usepackage{wrapfig}
\usepackage{enumitem}
\usepackage[table,xcdraw]{xcolor} 
\usepackage{pifont}
\usepackage{subcaption}
\usepackage[most]{tcolorbox}

\newtheorem{definition}{Definition}

\newtcolorbox{mydefbox}{colback=gray!10,
  boxrule=0pt, leftrule=0pt, rightrule=0pt, toprule=0pt, bottomrule=0pt,breakable, enhanced}

\definecolor{cvprblue}{rgb}{0.21,0.49,0.74}
\usepackage[pagebackref,breaklinks,colorlinks,allcolors=cvprblue]{hyperref}

\def\paperID{} 
\def\confName{CVPR}
\def\confYear{2026}

\title{Decoupling Stability and Plasticity for Multi-Modal Test-Time Adaptation}

\author{Yongbo He \quad Zirun Guo \quad Tao Jin\thanks{Corresponding author.}\\
Zhejiang University\\
{\tt\small reheybo@gmail.com, jint\_zju@zju.edu.cn}
}

\begin{document}

\maketitle

\begin{abstract}
    Adapting pretrained multi-modal models to evolving test-time distributions, known as multi-modal test-time adaptation, presents a significant challenge. Existing methods frequently encounter \textbf{negative transfer in the unbiased modality} and \textbf{catastrophic forgetting in the biased modality}. To address these challenges, we propose \textbf{D}ecoupling \textbf{A}daptation for \textbf{S}tability and \textbf{P}lasticity (\textbf{DASP}), a novel \textbf{diagnose-then-mitigate} framework. Our analysis reveals a critical discrepancy within the unified latent space: the biased modality exhibits substantially higher interdimensional redundancy (\ie strong correlations across feature dimensions) compared to the unbiased modality. Leveraging this insight, DASP \textbf{identifies the biased modality} and \textbf{implements an asymmetric adaptation strategy}. This strategy employs a decoupled architecture where each modality-specific adapter is divided into stable and plastic components. The asymmetric mechanism works as follows: for the biased modality, which requires plasticity, the plastic component is activated and updated to capture domain-specific information, while the stable component remains fixed. Conversely, for the unbiased modality, which requires stability, the plastic component is bypassed, and the stable component is updated using KL regularization to prevent negative transfer. This asymmetric design enables the model to adapt flexibly to new domains while preserving generalizable knowledge. Comprehensive evaluations on diverse multi-modal benchmarks demonstrate that DASP significantly outperforms state-of-the-art methods. Code is available at \url{https://github.com/he4cs/DASP}.
\end{abstract}    

\begin{figure}[t]
    \centering
    \includegraphics[width=\linewidth]{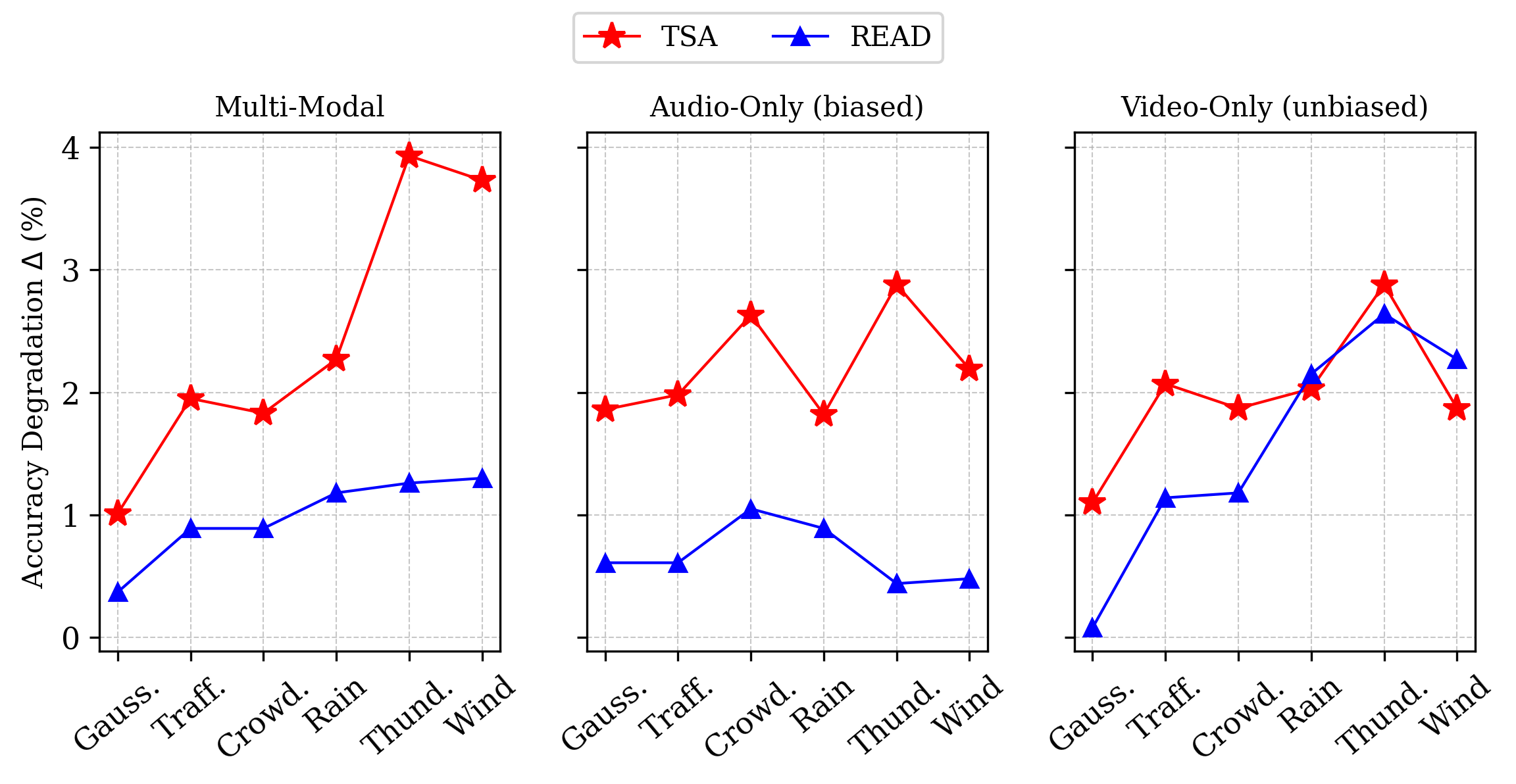}
    \caption{\textbf{Limitations in Multi-Modal TTA.} We evaluate changes in source domain performance during continual adaptation, measured as $\Delta=\mathrm{Acc}_\text{orignal}-\mathrm{Acc}_\text{adapted}$, for state-of-the-art methods (READ and TSA). Results indicate ongoing degradation in both multi-modal and uni-modal contexts. Performance drops in the biased modality are referred to as \textit{catastrophic forgetting}, while drops in the unbiased modality are considered \textit{negative transfer}.}
    \label{fig:limit}
    \vspace{-5mm}
\end{figure}

\section{Introduction}
\label{sec:intro}

Multi-modal models~\citep{clip,cggm,mars_sep,vif,visdoc,vismem,think3d} integrate and leverage complementary information from diverse sensors to improve perception and understanding. However, these approaches remain vulnerable in non-stationary environments. In open-world scenarios, test samples are subject to natural variations and unforeseen corruptions (\ie \textit{distribution shifts}), which arise from unpredictable factors such as weather changes and sensor degradation~\citep{robustness}. Such distribution shifts in multi-modal data present challenges, such as a decrease in discriminative information and misalignment between different modalities. Consequently, static pre-trained models experience notable performance degradation. To address this, test-time adaptation (TTA) has recently emerged as a promising method to convert static models into adaptive systems. By updating model parameters online, TTA enables models to adapt to distribution shifts during testing without requiring access to source data.

Recent TTA methods have demonstrated effectiveness primarily through entropy minimization techniques, as shown in studies like Tent~\citep{tent}, EATA~\citep{eata}, and SAR~\citep{sar}. Building on these foundations, researchers have investigated advanced strategies for multi-modal variants, including self-adaptive fusion~\citep{read}, reliable sample selection~\citep{sumi}, and selective adaptation~\citep{tsa}. Despite these advancements, two significant limitations remain, as illustrated in~\cref{fig:limit}: \ding{182} Modality-agnostic strategies, which adapt all modalities indiscriminately, can cause \textit{negative transfer} to modalities that are already well-aligned. This occurs because the model overfits noise in the unsupervised signal, thereby degrading performance in unbiased modalities. \ding{183} Continuous parameter updates may result in \textit{catastrophic forgetting}. While adapting to the target distribution requires substantial parameter changes, these domain-specific updates risk erasing knowledge acquired from the source domain. Overall, current methods struggle to balance stability (maintaining source-domain performance) and plasticity (effectively adapting to the target domain). This fundamental trade-off is referred to as the \textbf{stability-plasticity dilemma}.

\begin{figure}[t]
    \centering
    \includegraphics[width=\linewidth]{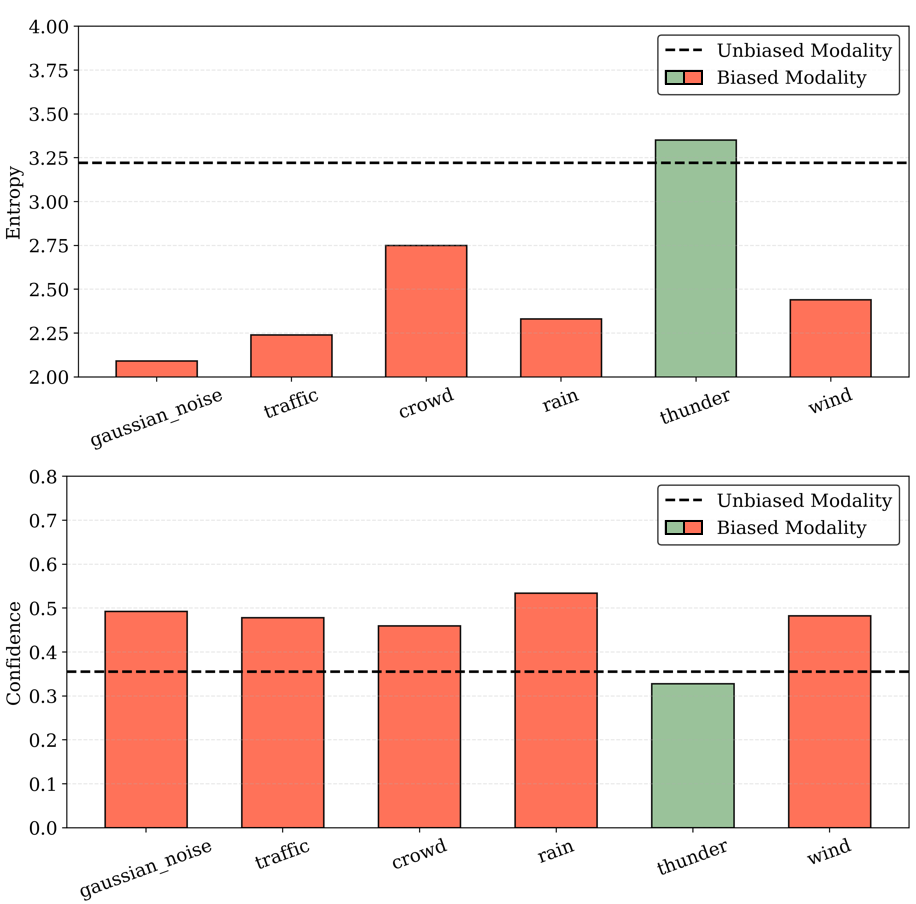}
    \caption{\textbf{Entropy and confidence statistics} on the VGGSound-C with corrupted audio modality. Since audio serves as the dominant modality in this dataset, it continues to display lower entropy and greater confidence, even in the presence of distribution shifts.}
    \label{fig:ent_conf}
    \vspace{-3mm}
\end{figure}

\begin{figure*}[t]
    \centering
    \includegraphics[width=\linewidth]{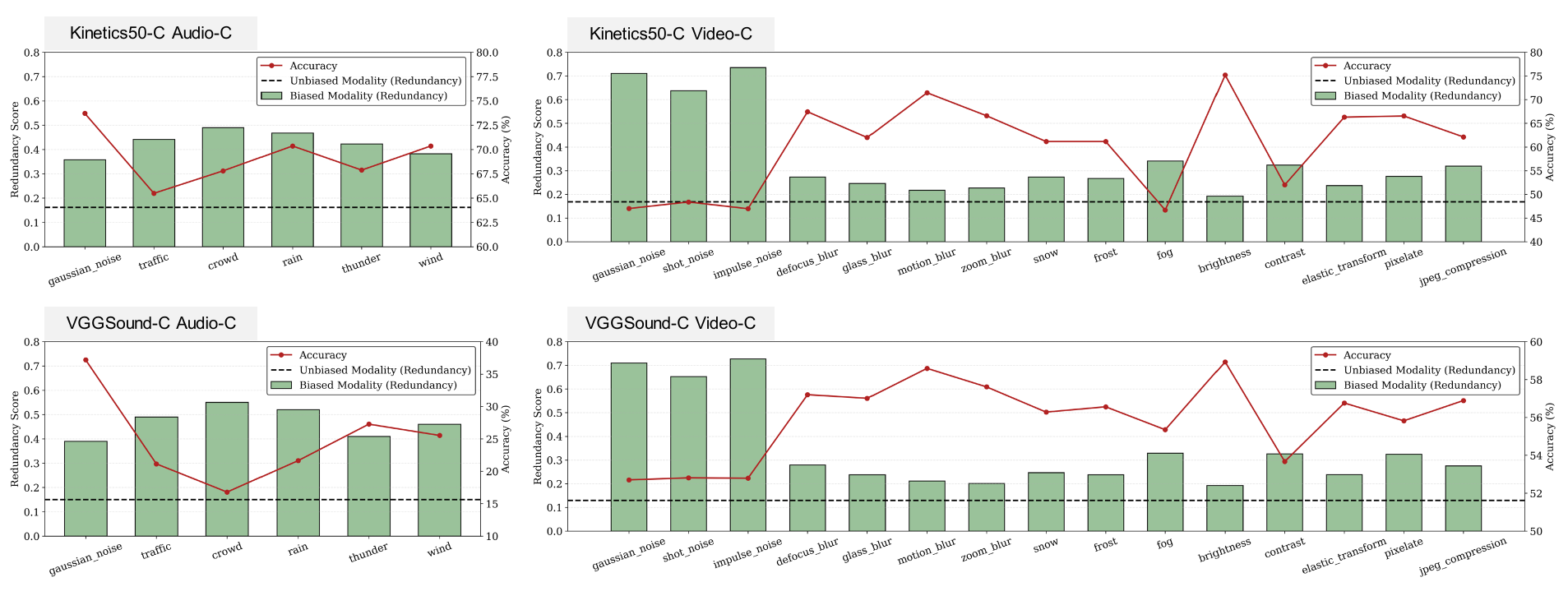}
    \caption{\textbf{Redundancy statistics} on Kinetics50-C and VGGSound-C. The corrupted modality demonstrates increased redundancy in feature embeddings. Furthermore, the results underscore a significant correlation between redundancy and accuracy.}
    \label{fig:redundancy}
    \vspace{-5mm}
\end{figure*}

To address these limitations, we propose a novel approach that \textbf{D}ecouples multi-modal \textbf{A}daptation for \textbf{S}tability and \textbf{P}lasticity (\textbf{DASP}). This enables the model to selectively prioritize stability or plasticity for each modality according to its distribution. Our approach employs a diagnose-then-mitigate framework: \ding{182} \textbf{Diagnosing via Redundancy Score.} In this phase, we identify modality bias. As illustrated in~\cref{fig:ent_conf}, commonly used metrics such as entropy or confidence are often unreliable for detecting distribution shifts without access to source domain statistics, due to inherent information discrepancies between dominant and auxiliary modalities. Instead, we analyze feature representations at the fusion layer: when modality features are projected into a unified space, the resulting embeddings exhibit comparable interdimensional correlations. Notably, these correlations increase significantly under distribution shifts. To measure this effect, we define a redundancy score and validate its effectiveness across multiple multi-modal datasets, as shown in~\cref{fig:redundancy}. \ding{183} \textbf{Mitigating via Asymmetric Adaptation.} Based on this diagnosis, we introduce asymmetric adaptation. Each modality-specific adapter consists of two components: stable and plastic adapters. Our method adapts the biased modality through its plastic adapter while keeping the stable adapter frozen. In contrast, for the unbiased modality, the plastic adapter is deactivated, and the stable adapter updates steadily under KL regularization. This mechanism externalizes the domain-specific parameters (\ie plastic adapter) while internalizing the domain-agnostic parameters (\ie stable adapter). Essentially, our method mitigates negative transfer in the unbiased modality through the stable adapter and avoids catastrophic forgetting by activating the plastic adapter when needed. Our contributions are summarized as follows:

\begin{itemize}[itemsep=0.1em]
    \item We highlight the stability-plasticity dilemma in multi-modal TTA and propose a novel approach, Decoupled Adaptation for Stability and Plasticity (DASP), which follows a diagnose-then-mitigate framework.
    \item We observe discrepancies in interdimensional correlations among biased and unbiased modalities. As a result, we define a redundancy score to identify modality bias.
    \item We design an asymmetric adaptation strategy for multi-modal models to address the limitations of negative transfer and catastrophic forgetting.
    \item We perform comprehensive experiments on Kinetics50-C and VGGSound-C, and DASP exhibits enhanced adaptivity and stability in comparison to existing methods.
\end{itemize}

\section{Related Work}
\label{sec:related}

\textbf{Test-Time Adaptation.} Prior studies address distribution shifts during training through domain generalization~\citep{cgt,mal}, enlarged training datasets~\citep{enlarged}, and diverse data augmentation strategies~\citep{augmix,selaug}. These approaches aim to anticipate potential test-time shifts by broadening the training distribution. However, exhaustively covering unseen shifts is infeasible, and such strategies often incur substantial computational costs. Test-Time Adaptation (TTA)~\citep{memo,ttn,foa,rotta} instead updates pre-trained models directly on changing data distributions at deployment. This setting relaxes the assumption that training and test distributions should match while not restricting or changing model training. Recently, entropy-based methods~\citep{tent,eata,sar,zerosiam} have demonstrated strong effectiveness in unsupervised TTA, as prediction entropy correlates closely with target-domain accuracy. Whereas earlier studies focused on individual fixed shifts, more recent works~\citep{cotta,vida,ecotta,becotta} have proposed approaches for continual adaptation over extended timescales, referred to as Continual Test-Time Adaptation (CTTA). In contrast to a single domain shift, CTTA is more vulnerable to catastrophic forgetting and the accumulation of errors. 

\smallskip
\noindent \textbf{Multi-Modal Test-Time Adaptation.} Multi-modal data improves perception in intelligent systems but remains vulnerable to distribution shifts across different modalities. Existing TTA methods, primarily developed for uni-modal tasks, inadequately address these complex shifts. In this context,~\citet{read} first highlighted the challenge of reliability bias in multi-modal TTA (MM-TTA) and proposed a solution that combines reliable fusion with robust adaptation. Subsequent research has focused on robust adaptation by integrating techniques such as attention bootstrapping~\citep{abpem}, smoothing the shift~\citep{sumi}, and partition-then-adapt~\citep{pta}. For multi-modal continual TTA (MM-CTTA) scenarios, MDAA~\citep{mdaa} mitigates catastrophic forgetting using analytic continual learning. Among previous works, TSA~\citep{tsa} is most relevant to our work, introducing selective adaptation via a routing module. However, this soft selection often lacks effectiveness and stability in unsupervised settings. In this work, we reveal that these multi-modal variants remain limited by negative transfer in unbiased modalities and forgetting in biased modalities. To address these challenges, we propose DASP, an asymmetric adaptation framework that decouples stability and plasticity, thereby effectively reducing both negative transfer and forgetting.

\section{Method}
\label{method}

\begin{figure*}[t]
    \centering
    \includegraphics[width=0.95\linewidth]{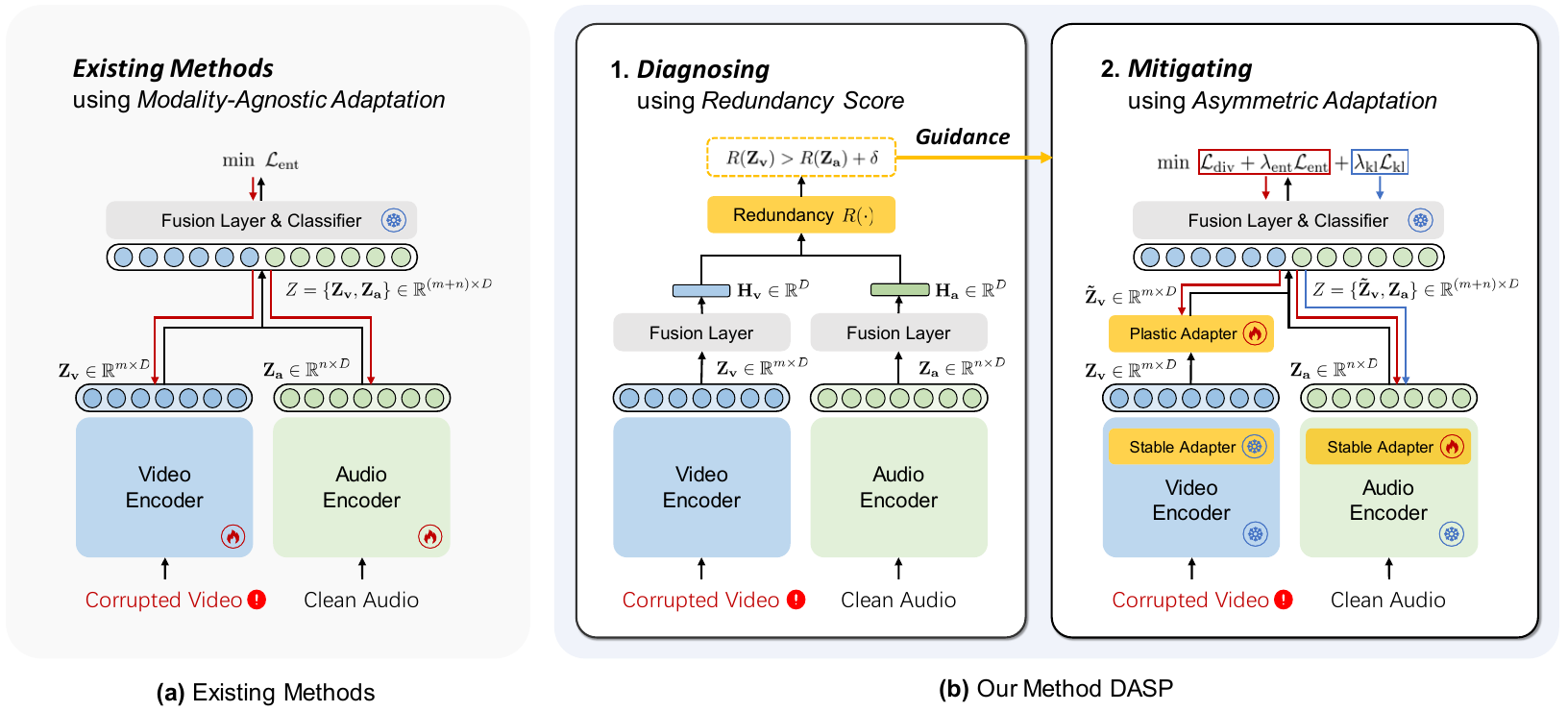}
    \caption{The overview of our proposed DASP features a diagnose-then-mitigate framework. It begins by diagnosing the biased modality through redundancy scores, followed by asymmetric adaptation that includes modality-specific updates guided by entropy minimization.}
    \label{fig:pipline}
    \vspace{-5mm}
\end{figure*}

In this section, we present a novel approach for multi-modal test-time adaptation (MM-TTA), as defined in Sec.~\ref{sec:definition}. Our method, called DASP (\cref{fig:pipline}), consists of two main components: (i) diagnosing via redundancy score (Sec.~\ref{sec:diagnosis}), and (ii) mitigation via asymmetric adaptation  (Sec.~\ref{sec:mitigation}).

\subsection{Problem Definition}
\label{sec:definition}

Without loss of generality, we formulate MM-TTA in the context of an audio-video classification task. Specifically, we employ a standard multi-modal architecture comprising modality-specific encoders and a unified fusion module, denoted as
$f = \{ f^a, f^v, f^u, f^c \},$
where $f^a$ and $f^v$ represent ViT-based encoders for the audio and video modalities, respectively, while $f^u$ and $f^c$ denote the fusion module and the classification head.
Domain adaptation aims to transfer a model from a source domain $P(\mathbf{x})$ to a target domain $Q(\mathbf{x})$, where $P(\mathbf{x})$ and $Q(\mathbf{x})$ exhibit a substantial distributional discrepancy. The base model $f_{\Theta}$, parameterized by $\Theta$, is first pretrained on a labeled dataset
$\mathcal{D}_{\text{source}} = \{(\mathbf{x}_i, y_i)\}_{i=1}^{N}$,
where each multi-modal sample $\mathbf{x}_i = \{x_i^a, x_i^v\} \sim P(\mathbf{x})$ consists of paired audio $x_i^a$ and video $x_i^v$ inputs. Although $f_{\Theta}$ achieves strong performance on in-distribution (ID) test samples drawn from $P(\mathbf{x})$, its generalization degrades significantly on out-of-distribution (OOD) samples 
$\mathcal{D}_{\text{target}} = \{\mathbf{x}_i\}_{i=1}^{M} \sim Q(\mathbf{x})$.

\subsection{Diagnosis via Redundancy Score}
\label{sec:diagnosis}

\noindent \textbf{Motivation.}  
Identifying biased modalities in MM-TTA without external supervision remains a significant challenge. Parametric methods~\citep{tsa} often struggle with reliable convergence due to the lack of consistent guidance, and an incorrect diagnosis can severely degrade adaptation performance. Building on the insight that distribution shifts manifest as changes in sample-level statistical properties, we propose a statistical indicator to effectively determine if a modality is corrupted by a distributional shift. Conventional diagnostic works~\citep{eata, sar} typically rely on single-modality metrics like entropy or confidence to detect shifts. However, in multi-modal systems, a dominant modality can inherently possess lower entropy and higher confidence than its auxiliary counterpart, even when facing a distribution shift (see ~\cref{fig:ent_conf}). This makes direct comparison across modalities unreliable for diagnosis. To overcome this limitation, we adopt the interdimensional redundancy score $R(\cdot)$ from $\mathrm{SAR^2}$~\citep{sar2}, a non-parametric metric that quantifies the internal correlation structure within feature representations.

\begin{mydefbox}
\begin{definition}[Interdimensional Redundancy]
Let $z_b = h(x_b) \in \mathbb{R}^D$ denote the feature representation of sample $x_b$ within a batch, and let $\mathbf{Z} = [z_1, \dots, z_B]$ denote the batch feature matrix with centroid 
$\mu_B = \frac{1}{B} \sum_{b=1}^B z_b$.

The interdimensional redundancy score is defined as
\begin{equation}
R(\mathbf{Z}) = \frac{1}{D(D-1)} \sum_{i \neq j} \mathcal{C}_{ij}^2,
\end{equation}
where $\mathcal{C}$ is the normalized covariance coefficient matrix given by
\begin{equation}
\mathcal{C} = \frac{1}{B} \cdot 
\frac{(\mathbf{Z} - \mu_B)^\top (\mathbf{Z} - \mu_B)}
{\sigma_B(\mathbf{Z})^2},
\end{equation}
and $\sigma_B(\mathbf{Z})$ denotes the per-dimension standard deviation.
\end{definition}
\end{mydefbox}

\smallskip
\noindent \textbf{Theoretical Analysis.}
Theoretically, a robust and generalizable latent space $\mathbf{Z}$ should exhibit linear disentanglement (\ie decorrelation), where each dimension captures an independent factor of variation. This objective is a tractable proxy for statistical independence, and it is widely leveraged in modern self-supervised learning to enforce dimensional diversity. The $R(\mathbf{Z})$ directly quantifies this linear dependency, where a maximally decorrelated representation would yield $R(\mathbf{Z}) \approx 0$. We posit that distribution shifts induce a structural degradation of the feature manifold. When an encoder $f^m$ processes a biased modality, it fails to maintain this decorrelation property. The feature dimensions become spuriously correlated (\eg all dimensions reacting identically to domain-specific noise), capturing highly redundant information instead of independent semantic factors. This degradation is reliably quantified by a significant elevation in $R(\mathbf{Z}^m)$, establishing the score as a measure of the feature representation's structural integrity.

\smallskip
\noindent \textbf{Observation.}
Let $\mathcal{M} = \{a, v\}$ denote the modality set for audio and video. Since the multi-modal features $z^m=f^u(f^m(x^m))_,m\in \mathcal{M}$ are projected into a shared latent space after the fusion module $f^u$, redundancy scores computed across modalities are directly comparable. As illustrated in~\cref{fig:redundancy}, redundancy correlates strongly with model accuracy, \ie biased modalities exhibit a noticeable increase in $R(\mathbf{Z})$. Therefore, we propose that redundancy discrepancy across modalities can serve as a robust indicator for identifying biased modalities.

\smallskip
\noindent \textbf{Rule-based Diagnosis.}
For a mini-batch $\mathcal{B}$ of target-domain samples, the projected modality-specific representations are
$\mathbf{Z}^{m} = [z_b^{m}]_{b \in \mathcal{B}} \in \mathbb{R}^{B \times D}, m \in \mathcal{M}$,
with corresponding redundancy scores
$r^{m} = R(\mathbf{Z}^{m})$.
The relative redundancy of modality $m$ is computed as
\begin{align}
    \Delta^{m} = r^{m} - \min_{n \in \mathcal{M}} r^{n}.
\end{align}
Given a predefined threshold $\delta > 0$, the set of biased modalities is identified as
\begin{align}
    \mathcal{G} = \{\, m \in \mathcal{M} \mid \Delta^{m} \geq \delta \,\},
\end{align}
and the unbiased modality set is $\mathcal{M} \setminus \mathcal{G}$.  
This rule provides an interpretable and deterministic mechanism for identifying modality bias.

\subsection{Mitigation via Asymmetric Adaptation}
\label{sec:mitigation}

\noindent \textbf{Motivation.}
Existing methods face two notable limitations in practical applications: \ding{182} Distribution shifts usually do not affect every modality in a sample. However, current modality-agnostic methods treat all modalities as equally important for adaptation. This assumption can cause negative transfer, where noisy adaptation signals harm the performance of well-aligned and unbiased modalities. \ding{183} Distributional shifts do not persist within a modality. Existing methods incorporate domain-specific parameters directly into the source model, leading to performance degradation in the source domain (where no shift exists), a phenomenon known as catastrophic forgetting. These methods struggle to perform well in both source and target domains. We attribute this problem to the stability-plasticity dilemma. 

To address this issue, we reconsider the role of various modalities in MM-TTA. Our analysis reveals distinct requirements: Biased modalities (\ie $m \in \mathcal{G}$) need significant parameter changes to adapt to the target distribution, meaning they require flexibility during adaptation. In contrast, unbiased modalities (\ie $m \in \mathcal{M} \setminus \mathcal{G}$) should remain stable and anchored to the source domain. Their complementary nature helps reduce information gaps caused by modality bias. Based on these insights, we propose an asymmetric adaptation method to decouple the stability and plasticity.

\smallskip
\noindent \textbf{Asymmetric Adaptation.} 
Our approach divides the modality-specific adapter $\Phi^m$ into two components: a stable adapter $\phi^m_{\text{s}}$ and a plastic adapter $\phi^m_{\text{p}}$. The plastic adapter is specifically activated to adapt distribution shifts, thereby isolating the domain-specific parameters and preserving knowledge embedded in the source domain model. The stable adapter acts as an internal module, continuously improving the model's ability to generalize. To emphasize their different roles, we implement the stable and plastic adapters with low-rank and high-rank structures, respectively. The low-rank design limits capacity to encourage domain-agnostic generalization, while the high-rank design provides enough parameters to capture complex, domain-specific information. Let $z^{m} = f^{m}(x^{m})$ represent the original feature. The adapted feature $\tilde{z}^{m}$ is computed as
\begin{align}
    \tilde{z}^{m} = \phi^m_{\text{p}}(\phi^m_{\text{s}}(z^{m})), \quad m \in \mathcal{G}, \\
    \tilde{z}^{m} = \phi^m_{\text{s}}(z^{m}), \quad m \in \mathcal{M} \setminus \mathcal{G},
\end{align}
During the adaptation phase, we employ an asymmetric adaptation strategy. For $m \in \mathcal{G}$, we update $\phi^m_{\text{p}}$ while freezing $\phi^m_{\text{s}}$ (\ie preventing gradient flow) to efficiently align the shifted feature space. Conversely, for $m \in \mathcal{M} \setminus \mathcal{G}$, we update $\phi^m_{\text{s}}$ while keeping $\phi^m_{\text{p}}$ inactive. This dynamic control ensures that parameters associated with the general representation ($\phi^m_{\text{s}}$) are only modified under stable conditions. Additionally, to preserve the stability of $\phi^m_{\text{s}}$, we incorporate a Kullback-Leibler (KL) penalty term, denoted as $\mathcal{L}_{\text{kl}}$, defined as follows:
\begin{align}
\label{eq:kl}
&\mathcal{L}_{\text{kl}}
= \mathbf{D}_\mathrm{KL}\!\left(p_{\text{tgt}}^{m} \,\middle\|\, p_{\text{src}}^{m}\right), \quad m \in \mathcal{M} \setminus \mathcal{G}, \\
&p_{\text{src}}^{m} = f^c(f^u(z^{m})), \quad p_{\text{tgt}}^{m} = f^c(f^u(\tilde{z}^{m})).
\end{align}
This term regularizes the distance from the source model, which helps to reduce negative transfer.

\smallskip
\noindent \textbf{Optimization.} To adapt the model without access to target domain annotations, we employ a standard unsupervised objective widely adopted in TTA. The entropy minimization loss is defined as:
\begin{align}
\label{eq:ent}
    \mathcal{L}_{\text{ent}}
    = - \sum_{y \in \mathcal{C}} f(y|\mathbf{x}) 
      \log f(y|\mathbf{x}),
\end{align}
where $\mathcal{C}$ denotes the label space.  
Following prior work~\citep{shot}, we further include a diversity-promoting regularization:
\begin{align}
\label{eq:div}
    \mathcal{L}_{\text{div}}
    = \sum_{y \in \mathcal{C}} \hat{p}_y \log \hat{p}_y,
\end{align}
where $\hat{p}_y = \frac{1}{B} \sum_{i=1}^{B} f(y|\mathbf{x}_i)$ represents the batch-averaged class probability for a mini-batch of size $B$.

By combining Eqs.~\ref{eq:kl}–\ref{eq:div}, our overall loss function for DASP can be written as:
\begin{align}
    \mathcal{L}_{\text{total}}
    = \mathcal{L}_{\text{div}} + \lambda_{\text{ent}} \mathcal{L}_{\text{ent}} + \lambda_{\text{kl}} \mathcal{L}_{\text{kl}},
\end{align}
where $\lambda_{\text{ent}}$ and $\lambda_{\text{kl}}$ are the balancing hyperparameters that control the trade-off among entropy minimization, diversity, and KL regularization. This combination encourages confident yet diverse predictions, enabling stable and effective adaptation in the absence of labeled data.

\begin{table*}[t]
    \caption{\textbf{Episodic Adaptation.} Comparison with SOTA methods on Kinetics50-C with \textbf{video corruptions} regarding \textbf{Accuracy (\%, $\uparrow$)}.}
    \label{exp:ks_v}
    \centering
    \begin{threeparttable}
    \large
    \renewcommand\arraystretch{1.2}
    \resizebox{\textwidth}{!}{
    \begin{tabular}{lcccccccccccccccc}
        \toprule
        \multicolumn{1}{c}{} & \multicolumn{3}{c}{Noise} & \multicolumn{4}{c}{Blur} & \multicolumn{4}{c}{Weather} & \multicolumn{4}{c}{Digital} & \multicolumn{1}{c}{} \\ 
        \cmidrule(lr){2-4} \cmidrule(lr){5-8} \cmidrule(lr){9-12} \cmidrule(lr){13-16}
        \multicolumn{1}{c}{Method} & Gauss. & Shot & Impul. & Defoc. & Glass & Mot. & Zoom & Snow & Frost & Fog & Brit. & Contr. & Elas. & Pix. & JPEG & \cellcolor{green!10} Avg. \\ 
        \midrule
        Source & 46.8 & 48.0 & 46.9 & 67.5 & 62.2 & 70.8 & 66.7 & 61.6 & 60.3 & 46.7 & 75.2 & 52.1 & 65.7 & 66.5 & 61.9 & \cellcolor{green!10} 59.9 \\
        $\bullet$ Tent \texttt{(ICLR'21)} & 46.3 & 47.0 & 46.3 & 67.2 & 62.5 & 71.0 & 67.6 & 63.1 & 61.1 & 34.9 & 75.4 & 51.6 & 66.8 & 67.2 & 62.7 & \cellcolor{green!10} 59.4 \\
        $\bullet$ EATA \texttt{(ICML'22)} & 46.8 & 47.6 & 47.1 & 67.6 & 62.7 & 70.7 & 66.7 & 62.2 & 63.0 & 49.2 & 75.2 & 52.4 & 65.9 & 67.0 & 59.4 & \cellcolor{green!10} 60.1 \\
        $\bullet$ SAR \texttt{(ICLR'23)} & 46.7 & 47.4 & 46.8 & 67.0 & 61.9 & 70.4 & 66.4 & 61.4 & 60.6 & 46.0 & 75.2 & 52.1 & 65.7 & 66.4 & 62.0 & \cellcolor{green!10} 59.8 \\
        $\bullet$ READ \texttt{(ICLR'24)} & 49.4 & 49.7 & 49.0 & 68.0 & 65.1 & 71.2 & 69.0 & 64.5 & 64.4 & 57.4 & \textbf{75.5} & 53.6 & 68.3 & 68.0 & 65.1 & \cellcolor{green!10} 62.5 \\
        $\bullet$ TSA \texttt{(ICML'25)} & 50.7 & 51.1 & 50.4 & 67.9 & 67.1 & 71.7 & 69.2 & 65.5 & 66.2 & 61.3 & 75.2 & 56.2 & 69.5 & 68.8 & 66.6 & \cellcolor{green!10} 63.8 \\
        \rowcolor{blue!10}
        \textbf{$\bullet$ Ours} & \textbf{50.8} & \textbf{51.6} & \textbf{50.7} & \textbf{70.2} & \textbf{69.3} & \textbf{72.3} & \textbf{71.3} & \textbf{66.1} & \textbf{68.2} & \textbf{63.5} & 75.2 & \textbf{58.1} & \textbf{71.2} & \textbf{70.5} & \textbf{68.6} & \textbf{65.2} \\
        \bottomrule
    \end{tabular}
    }
    \end{threeparttable}
\end{table*}

\begin{table*}[!h]
    \centering
    \caption{\textbf{Episodic Adaptation.} Comparison with SOTA methods on Kinetics50-C (left) and VGGSound-C (right) with \textbf{audio corruptions} regarding \textbf{Accuracy (\%, $\uparrow$)}.}
    \label{exp:vgg_ks_a}
    \resizebox{\linewidth}{!}{
    \renewcommand\arraystretch{1.2}
    \begin{tabular}{lccccccc|cccccccc}
    \toprule
    \multicolumn{1}{c}{} & \multicolumn{3}{c}{Noise} & \multicolumn{3}{c}{Weather} & & \multicolumn{3}{c}{Noise} & \multicolumn{3}{c}{Weather} &  \\ 
    \cmidrule(lr){2-4} \cmidrule(lr){5-7} \cmidrule(lr){9-11} \cmidrule(lr){12-14} 
    \multicolumn{1}{c}{Method} & Gauss. & Traff. & Crowd. & Rain & Thund. & Wind & \cellcolor{green!10} Avg. & Gauss. & Traff. & Crowd. & Rain & Thund. & Wind & \cellcolor{green!10} Avg. \\ 
    \midrule
    Source 
        & 73.7 & 65.5 & 67.9 & 70.3 & 67.9 & 70.3 & \cellcolor{green!10} 69.3 
        & 37.0 & 25.5 & 16.8 & 21.6 & 27.3 & 25.5 & \cellcolor{green!10} 25.6 \\
    $\bullet$ Tent \texttt{(ICLR'21)}
        & 73.9 & 67.4 & 69.2 & 70.4 & 66.5 & 70.5 & \cellcolor{green!10} 69.6 
        & 10.6 & 2.6 & 1.8 & 2.8 & 5.3 & 4.1 & \cellcolor{green!10} 4.5 \\
    $\bullet$ EATA \texttt{(ICML'22)}
        & 73.7 & 66.1 & 68.5 & 70.2 & 67.3 & 70.2 & \cellcolor{green!10} 69.4 
        & 39.2 & 26.1 & 22.9 & 26.0 & 31.7 & 30.4 & \cellcolor{green!10} 29.4 \\
    $\bullet$ SAR \texttt{(ICLR'23)}
        & 73.7 & 65.4 & 68.2 & 69.9 & 67.0 & 70.2 & \cellcolor{green!10} 69.1 
        & 37.4 & 9.5  & 9.0  & 12.6 & 23.7 & 20.1 & \cellcolor{green!10} 20.4 \\
    $\bullet$ READ \texttt{(ICLR'24)}
        & 74.1 & 69.0 & 69.7 & 71.1 & 71.8 & 70.7 & \cellcolor{green!10} 71.1 
        & 40.4 & 28.9 & 26.0 & 30.9 & 36.0 & 32.6 & \cellcolor{green!10} 32.4 \\
    $\bullet$ TSA \texttt{(ICML'25)}
        & 74.3 & 69.3 & 70.4 & 71.2 & 72.3 & 70.6 & \cellcolor{green!10} 71.3 
        & 41.6 & 31.8 & 30.5 & 32.7 & 38.7 & 32.9 & \cellcolor{green!10} 34.7 \\
    \rowcolor{blue!10} 
    \textbf{$\bullet$ Ours} 
        & \textbf{75.7} & \textbf{71.5} & \textbf{71.8} & \textbf{72.2} & \textbf{74.5} & \textbf{71.9} & \textbf{72.9}
        & \textbf{43.4} & \textbf{37.5} & \textbf{38.4} & \textbf{36.3} & \textbf{44.6} & \textbf{37.9} & \textbf{39.7} \\
    \bottomrule
    \end{tabular}
    }
    \vspace{-3mm}
\end{table*}

\begin{table*}[t]
    \caption{\textbf{Continual Adaptation.} Comparison with SOTA methods on Kinetics50-C with \textbf{video corruptions} regarding \textbf{Accuracy (\%, $\uparrow$)}.}
    \label{exp:ks_v_continual}
    \centering
    \begin{threeparttable}
    \large
    \renewcommand\arraystretch{1.2}
    \resizebox{\textwidth}{!}{
    \begin{tabular}{lcccccccccccccccc}
        \toprule
        \multicolumn{1}{c}{} & \multicolumn{3}{c}{Noise} & \multicolumn{4}{c}{Blur} & \multicolumn{4}{c}{Weather} & \multicolumn{4}{c}{Digital} & \multicolumn{1}{c}{} \\ 
        \cmidrule(lr){2-4} \cmidrule(lr){5-8} \cmidrule(lr){9-12} \cmidrule(lr){13-16}
        \multicolumn{1}{c}{Method} & Gauss. & Shot & Impul. & Defoc. & Glass & Mot. & Zoom & Snow & Frost & Fog & Brit. & Contr. & Elas. & Pix. & JPEG & \cellcolor{green!10} Avg. \\ 
        \midrule
        \multicolumn{1}{c}{} &
        \multicolumn{15}{c}{$t \; \xrightarrow{\hspace{18cm}}$} \\ 
        \midrule
        Source & 46.8 & 48.0 & 46.9 & 67.5 & 62.2 & 70.8 & 66.7 & 61.6 & 60.3 & 46.7 & 75.2 & 52.1 & 65.7 & 66.5 & 61.9 & \cellcolor{green!10} 59.9 \\
        $\bullet$ Tent \texttt{(ICLR'21)} & 46.2 & 45.8 & 43.1 & 62.5 & 61.9 & 65.6 & 64.1 & 55.0 & 56.9 & 40.1 & 40.4 & 23.6 & 14.6 & 4.5 & 3.4 & \cellcolor{green!10} 41.8 \\
        $\bullet$ EATA \texttt{(ICML'22)} & 47.0 & 47.9 & 47.0 & 67.4 & 63.3 & \textbf{71.6} & 68.5 & 61.8 & 65.0 & 59.3 & 74.2 & 53.0 & 69.9 & 64.9 & 61.9 & \cellcolor{green!10} 61.5 \\
        $\bullet$ SAR \texttt{(ICLR'23)} & 47.0 & 48.0 & 46.8 & 66.2 & 62.2 & 70.7 & 66.8 & 60.2 & 60.6 & 49.8 & 74.9 & 51.0 & 66.5 & 65.0 & 61.1 & \cellcolor{green!10} 59.8 \\
        $\bullet$ READ \texttt{(ICLR'24)} & 49.9 & 51.2 & 52.0 & 68.6 & 66.6 & 69.3 & 67.6 & 61.9 & 63.9 & 59.0 & 69.1 & 52.5 & 65.3 & 60.3 & 59.3 & \cellcolor{green!10} 61.1 \\
        $\bullet$ TSA \texttt{(ICML'25)} & 50.7 & 52.3 & 52.7 & 67.5 & 67.2 & 68.7 & 67.8 & 62.7 & 65.9 & 59.9 & 71.3 & 54.3 & 67.0 & 64.8 & 62.9 & \cellcolor{green!10} 62.4 \\
        \rowcolor{blue!10} 
        \textbf{$\bullet$ Ours} & \textbf{50.8} & \textbf{52.4} & \textbf{52.0} & \textbf{69.2} & \textbf{70.4} & 71.5 & \textbf{71.1} & \textbf{65.3} & \textbf{68.6} & \textbf{66.3} & \textbf{75.2} & \textbf{57.7} & \textbf{71.1} & \textbf{70.2} & \textbf{67.8} & \textbf{65.3} \\ 
        \bottomrule
    \end{tabular}
    }
    \end{threeparttable}
\end{table*}

\begin{table*}[!h]
    \centering
    \caption{\textbf{Continual Adaptation.} Comparison with SOTA methods on Kinetics50-C (left) and VGGSound-C (right) with \textbf{audio corruptions} regarding \textbf{Accuracy (\%, $\uparrow$)}.}
    \label{exp:vgg_ks_a_continual}
    \resizebox{\linewidth}{!}{
    \renewcommand\arraystretch{1.2}
    \begin{tabular}{lccccccc|cccccccc}
    \toprule
    \multicolumn{1}{c}{} & \multicolumn{3}{c}{Noise} & \multicolumn{3}{c}{Weather} & & \multicolumn{3}{c}{Noise} & \multicolumn{3}{c}{Weather} &  \\ 
    \cmidrule(lr){2-4} \cmidrule(lr){5-7} \cmidrule(lr){9-11} \cmidrule(lr){12-14} 
    \multicolumn{1}{c}{Method} & Gauss. & Traff. & Crowd. & Rain & Thund. & Wind & \cellcolor{green!10} Avg. & Gauss. & Traff. & Crowd. & Rain & Thund. & Wind & \cellcolor{green!10} Avg. \\ 
    \midrule
    \multicolumn{1}{c}{} &
    \multicolumn{6}{c}{$t \; \xrightarrow{\hspace{6cm}}$} & 
    \multicolumn{1}{c}{} &
    \multicolumn{6}{c}{$t \; \xrightarrow{\hspace{6cm}}$} \\ 
    \midrule
    Source 
        & 73.7 & 65.5 & 67.9 & 70.3 & 67.9 & 70.3 & \cellcolor{green!10} 69.3 
        & 37.0 & 25.5 & 16.8 & 21.6 & 27.3 & 25.5 & \cellcolor{green!10} 25.6 \\
    $\bullet$ Tent \texttt{(ICLR'21)}
        & 73.9 & 67.9 & 71.1 & 70.1 & 71.7 & 70.6 & \cellcolor{green!10} 70.9 
        & 10.7 & 1.2 & 0.4 & 0.3 & 0.5 & 0.3 & \cellcolor{green!10} 2.3 \\
    $\bullet$ EATA \texttt{(ICML'22)}
        & 73.8 & 67.6 & 70.2 & 70.5 & 70.3 & 70.2 & \cellcolor{green!10} 70.4 
        & 40.4 & 33.0 & 35.5 & 34.9 & 43.6 & 37.1 & \cellcolor{green!10} 37.4 \\
    $\bullet$ SAR \texttt{(ICLR'23)}
        & 73.8 & 66.1 & 68.4 & 70.0 & 69.0 & 70.0 & \cellcolor{green!10} 69.6 
        & 37.9 & 8.9 & 7.4 & 15.1 & 13.5 & 18.2 & \cellcolor{green!10} 16.8 \\
    $\bullet$ READ \texttt{(ICLR'24)}
        & 74.2 & 68.5 & 70.4 & 70.1 & 71.7 & 69.2 & \cellcolor{green!10} 70.7 
        & 39.4 & 22.9 & 17.8 & 15.6 & 20.1 & 15.0 & \cellcolor{green!10} 21.8 \\
    $\bullet$ TSA \texttt{(ICML'25)}
        & 74.3 & 68.3 & 71.2 & 69.4 & 70.6 & 67.6 & \cellcolor{green!10} 70.2
        & 41.6 & 32.2 & 32.3 & 31.8 & 39.6 & 33.5 & \cellcolor{green!10} 35.2 \\
    \rowcolor{blue!10} 
    \textbf{$\bullet$ Ours} 
        & \textbf{75.7} & \textbf{70.5} & \textbf{72.6} & \textbf{71.2} & \textbf{73.6} & \textbf{71.2} & \textbf{72.5}
        & \textbf{43.4} & \textbf{36.8} & \textbf{38.5} & \textbf{36.6} & \textbf{44.2} & \textbf{38.3} & \textbf{39.6} \\
    \bottomrule
    \end{tabular}
    }
\end{table*}

\section{Experiments}

In this section, we conduct extensive experiments to answer the following research questions:
\begin{itemize}
    \item How does DASP compare to baseline methods in terms of episodic adaptation?
    \item Can DASP facilitate continual adaptation and mitigate catastrophic forgetting in scenarios involving both uni-modal and interleaved modality corruption?
    \item How do the components of DASP contribute to its performance, and how do hyperparameters affect its results?
\end{itemize}

\begin{table}[t]
    \caption{\textbf{Ablation Study.} Analysis of component contributions in our method on Kinetics50-C regarding \textbf{Accuracy (\%, $\uparrow$)}.}
    \label{exp:ablation}
    \centering
    \begin{threeparttable}
    \renewcommand\arraystretch{1.2}
    \resizebox{\linewidth}{!}{
    \begin{tabular}{lcccccc}
        \toprule
        \multicolumn{1}{c}{Method} & Video-C & Audio-C & Interleaved-C & \cellcolor{green!10} Avg. \\
        \midrule
        Source & 59.9 & 69.3 & 62.6 & \cellcolor{green!10} 63.9 \\
        $\bullet$ Ours & 65.3 & 72.5 & 67.3 & \cellcolor{green!10} 68.4 \\
        \quad w/o stable adapter & 64.6 & 72.3 & 65.2 & \cellcolor{green!10} 67.4 \\
        \quad w/o plastic adapter & 62.2 & 71.4 & 64.4 & \cellcolor{green!10} 66.0 \\
        \quad w/o asym. adaptation & 54.3 & 71.5 & 61.8 & \cellcolor{green!10} 62.5 \\
        \quad w/ asym. adaptation (opposite) & 52.3 & 71.8 & 61.5 & \cellcolor{green!10} 61.9 \\
        \bottomrule
    \end{tabular}
    }
    \end{threeparttable}
    \vspace{-5mm}
\end{table}

\begin{table*}[t]
    \caption{\textbf{Continual Adaptation.} Comparison with SOTA methods on Kinetics50-C with \textbf{interleaved modality corruption} regarding \textbf{Accuracy (\%, $\uparrow$)}.}
    \label{exp:ks_interleaved}
    \centering
    \begin{threeparttable}
    \LARGE
    \renewcommand\arraystretch{1.2}
    \resizebox{\textwidth}{!}{
    \begin{tabular}{lcccccccccccccccccccccc}
        \toprule
        \multicolumn{1}{c}{Method} & 
        \multicolumn{2}{c}{V-C} & \multicolumn{1}{c}{A-C} & \multicolumn{2}{c}{V-C} & \multicolumn{1}{c}{A-C} & \multicolumn{2}{c}{V-C} & \multicolumn{1}{c}{A-C} & \multicolumn{3}{c}{V-C} & \multicolumn{1}{c}{A-C} & \multicolumn{2}{c}{V-C} & \multicolumn{1}{c}{A-C} & \multicolumn{2}{c}{V-C} & \multicolumn{1}{c}{A-C} & \multicolumn{2}{c}{V-C} & \multicolumn{1}{c}{} \\
        \cmidrule(lr){2-3} \cmidrule(lr){4-4} \cmidrule(lr){5-6} \cmidrule(lr){7-7} \cmidrule(lr){8-9} \cmidrule(lr){10-10} \cmidrule(lr){11-13} \cmidrule(lr){14-14} \cmidrule(lr){15-16} \cmidrule(lr){17-17} \cmidrule(lr){18-19} \cmidrule(lr){20-20} \cmidrule(lr){21-22}
        & Gauss. & Shot & Gauss. & Impul. & Defoc. & Traff. & Glass & Mot. & Crowd. & Zoom & Snow & Frost & Rain & Fog & Brit. & Thund. & Contr. & Elas. & Wind & Pix. & JPEG & \cellcolor{green!10} Avg. \\
        \midrule
        \multicolumn{1}{c}{} & 
        \multicolumn{21}{c}{$t \; \xrightarrow{\hspace{35cm}}$} \\ 
        \midrule
        Source & 46.8 & 48.0 & 73.7 & 46.9 & 67.5 & 65.5 & 62.2 & 70.8 & 67.9 & 66.7 & 61.6 & 60.3 & 70.3 & 46.7 & 75.2 & 67.9 & 52.1 & 65.7 & 70.3 & 66.5 & 61.9 & \cellcolor{green!10} 62.6 \\
        $\bullet$ Tent \texttt{(ICLR'21)} & 46.3 & 45.8 & 73.2 & 42.9 & 61.6 & 66.1 & 62.5 & 66.6 & 68.5 & 64.1 & 53.7 & 56.9 & 65.5 & 45.6 & 59.1 & 23.6 & 32.6 & 29.7 & 22.5 & 5.1 & 3.6 & \cellcolor{green!10} 47.4 \\
        $\bullet$ EATA \texttt{(ICML'22)} & 47.0 & 47.9 & 74.0 & 47.2 & 67.2 & 66.8 & 65.0 & 70.8 & 69.7 & 68.7 & 59.7 & 64.1 & 69.5 & 58.4 & 73.1 & 68.7 & 53.0 & 69.1 & 69.1 & 64.9 & 61.8 & \cellcolor{green!10} 63.6 \\
        $\bullet$ SAR \texttt{(ICLR'23)} & 47.0 & 48.0 & 73.8 & 46.9 & 66.3 & 65.7 & 62.3 & 70.6 & 67.8 & 66.8 & 59.8 & 59.9 & 69.5 & 50.3 & 74.7 & 67.6 & 50.9 & 66.7 & 69.5 & 64.5 & 60.8 & \cellcolor{green!10} 62.4 \\

        $\bullet$ READ \texttt{(ICLR'24)} & 49.9 & 51.2 & 73.7 & 51.5 & 68.4 & 68.1 & 66.2 & 70.2 & 68.7 & 68.1 & 62.7 & 64.5 & 68.2 & 60.1 & 70.8 & 66.8 & 52.9 & 66.6 & 66.2 & 62.3 & 60.5 & \cellcolor{green!10} 63.7 \\
        $\bullet$ TSA \texttt{(ICML'25)} & 50.7 & 52.3 & 73.1 & 52.0 & 66.6 & 66.7 & 66.2 & 69.3 & 68.3 & 67.2 & 63.1 & 65.3 & 67.4 & 60.3 & 72.0 & 66.8 & 54.1 & 66.6 & 66.6 & 64.6 & 63.1 & \cellcolor{green!10} 63.9 \\
        \rowcolor{blue!10} 
        \textbf{$\bullet$ Ours} & \textbf{50.8} & \textbf{52.3} & \textbf{75.1} & \textbf{52.1} & \textbf{69.3} & \textbf{69.8} & \textbf{70.0} & \textbf{71.9} & \textbf{72.7} & \textbf{71.1} & \textbf{65.8} & \textbf{68.9} & \textbf{71.1} & \textbf{66.4} & \textbf{75.2} & \textbf{73.6} & \textbf{58.8} & \textbf{71.6} & \textbf{70.9} & \textbf{69.9} & \textbf{67.3} & \textbf{67.3} \\
        \bottomrule
    \end{tabular}
    }
    \end{threeparttable}
\end{table*}

\begin{table*}[t]
    \caption{\textbf{Continual Adaptation.} Comparison with SOTA methods on VGGSound-C with \textbf{interleaved modality corruption} regarding \textbf{Accuracy (\%, $\uparrow$)}.}
    \label{exp:vgg_interleaved}
    \centering
    \begin{threeparttable}
    \LARGE
    \renewcommand\arraystretch{1.2}
    \resizebox{\textwidth}{!}{
    \begin{tabular}{lcccccccccccccccccccccc}
        \toprule
        \multicolumn{1}{c}{Method} & 
        \multicolumn{2}{c}{V-C} & \multicolumn{1}{c}{A-C} & \multicolumn{2}{c}{V-C} & \multicolumn{1}{c}{A-C} & \multicolumn{2}{c}{V-C} & \multicolumn{1}{c}{A-C} & \multicolumn{3}{c}{V-C} & \multicolumn{1}{c}{A-C} & \multicolumn{2}{c}{V-C} & \multicolumn{1}{c}{A-C} & \multicolumn{2}{c}{V-C} & \multicolumn{1}{c}{A-C} & \multicolumn{2}{c}{V-C} & \multicolumn{1}{c}{} \\
        \cmidrule(lr){2-3} \cmidrule(lr){4-4} \cmidrule(lr){5-6} \cmidrule(lr){7-7} \cmidrule(lr){8-9} \cmidrule(lr){10-10} \cmidrule(lr){11-13} \cmidrule(lr){14-14} \cmidrule(lr){15-16} \cmidrule(lr){17-17} \cmidrule(lr){18-19} \cmidrule(lr){20-20} \cmidrule(lr){21-22}
        & Gauss. & Shot & Gauss. & Impul. & Defoc. & Traff. & Glass & Mot. & Crowd. & Zoom & Snow & Frost & Rain & Fog & Brit. & Thund. & Contr. & Elas. & Wind & Pix. & JPEG & \cellcolor{green!10} Avg. \\
        \midrule
        \multicolumn{1}{c}{} & 
        \multicolumn{21}{c}{$t \; \xrightarrow{\hspace{35cm}}$} \\ 
        \midrule
        Source & 52.8 & 52.7 & 37.0 & 52.7 & 57.2 & 25.5 & 57.2 & 58.7 & 16.8 & 57.0 & 56.4 & 56.6 & 21.6 & 55.6 & 58.0 & 27.3 & 53.7 & 56.9 & 25.5 & 55.8 & 56.9 & \cellcolor{green!10} 47.2 \\
        $\bullet$ Tent \texttt{(ICLR'21)} & 52.7 & 52.4 & 5.4 & 49.3 & 52.3 & 0.7 & 51.3 & 52.8 & 0.6 & 52.2 & 50.4 & 51.5 & 0.5 & 51.1 & 51.4 & 1.5 & 47.6 & 47.2 & 0.4 & 36.8 & 11.3 & \cellcolor{green!10} 34.3 \\
        $\bullet$ EATA \texttt{(ICML'22)} & 53.1 & 53.4 & 39.7 & 53.0 & 57.1 & 33.5 & 57.1 & 58.9 & 35.9 & 58.0 & 56.2 & 57.3 & 34.2 & 57.4 & 59.6 & 42.9 & 54.4 & 58.2 & 36.3 & 57.2 & \textbf{58.4} & \cellcolor{green!10} 51.0 \\
        $\bullet$ SAR \texttt{(ICLR'23)} & 52.9 & 53.0 & 35.5 & 52.7 & 57.0 & 11.2 & 56.9 & 58.4 & 10.3 & 57.8 & 56.2 & 56.9 & 13.5 & 55.6 & 57.9 & 18.6 & 53.3 & 56.3 & 21.3 & 54.6 & 55.6 & \cellcolor{green!10} 45.0 \\
        $\bullet$ READ \texttt{(ICLR'24)} & 53.6 & 54.0 & 35.0 & 53.9 & 57.6 & 20.7 & 57.3 & 58.4 & 18.6 & 57.9 & 56.4 & 56.9 & 15.6 & 56.7 & 58.0 & 22.0 & 55.0 & 56.8 & 14.8 & 56.1 & 56.0 & \cellcolor{green!10} 46.3 \\
        $\bullet$ TSA \texttt{(ICML'25)} & 53.5 & 54.1 & 40.3 & 53.9 & 57.3 & 0.8 & 5.1 & 0.4 & 0.3 & 0.4 & 0.3 & 0.3 & 0.3 & 0.3 & 0.3 & 0.3 & 0.3 & 0.3 & 0.3 & 0.4 & 0.3 & \cellcolor{green!10} 12.8 \\
        \rowcolor{blue!10} 
        \textbf{$\bullet$ Ours} & \textbf{54.0} & \textbf{54.9} & \textbf{42.5} & \textbf{55.0} & \textbf{58.3} & \textbf{36.7} & \textbf{58.7} & \textbf{59.9} & \textbf{39.0} & \textbf{59.5} & \textbf{57.5} & \textbf{58.2} & \textbf{36.2} & \textbf{58.3} & \textbf{60.0} & \textbf{44.6} & \textbf{54.5} & \textbf{58.5} & \textbf{38.2} & \textbf{57.7} & 58.3 & \textbf{52.5} \\
        \bottomrule
    \end{tabular}
     }
     \vspace{-3mm}
    \end{threeparttable}
\end{table*}

\subsection{Experimental Setups}

\textbf{Datasets and Models.}
Previous research~\citep{read} has constructed two audio-visual benchmark datasets, Kinetics~\citep{kinetic} and VGGSound~\citep{vggsound}, and introduced 15 types of video corruptions and 6 types of audio corruptions. The ViT~\citep{vit}-based CAV-MAE~\citep{cavmae} model serves as the source model, which is pre-trained on web-scale audio-visual data and fine-tuned on the training sets of the Kinetics50 and VGGSound datasets. 
In the experiments, the clean datasets are the source domain, and the corrupted datasets are the target domain. As a result, we obtain the Kinetics50-C and VGGSound-C benchmarks with either corrupted audio or corrupted video modalities. Each type of corruption has five levels of severity. In order to check the performance under the worst corruption case, we focus on testing with corrupted data of a high severity level.

\noindent \textbf{Compared Methods.}
To evaluate the effectiveness of the proposed method, we conduct comparative experiments against several state-of-the-art (SOTA) approaches in both uni-modal and multi-modal TTA. Specifically, Tent~\citep{tent}, EATA~\citep{eata}, and SAR~\citep{sar} represent entropy-based methods designed for uni-modal TTA, while READ~\citep{read} and TSA~\citep{tsa} are oriented to multi-modal TTA. READ introduces a self-adaptive attention-based fusion (SAF) mechanism, whereas TSA employs a selective adaptation scheme.

\noindent \textbf{Implementation Details.} For test-time adaptation, we update the model parameters using the Adam optimizer with a batch size of 64 and a learning rate of 0.0001 across all benchmarks. Furthermore, we keep the threshold $\delta$ at 0.05, while the loss coefficients $\lambda_{\text{ent}}$ and $\lambda_{\text{kl}}$ are fixed at 0.5 and 1.0, respectively, for all experiments.

\subsection{Main Results}

\noindent \textbf{Episodic Adaptation under Uni-Modal Corruption.} 
This evaluation examines model adaptation performance under uni-modal corruption using episodic adaptation. For each corruption type, model parameters are reset prior to adaptation to prevent interference from previous adaptations. Our proposed method shows marked superiority over existing approaches by effectively avoiding negative transfer in the unbiased modality. Notably, under audio corruption conditions, our method significantly outperforms the previous state-of-the-art (SOTA), achieving average performance improvements of 1.6\% on Kinetics50-C and 5.0\% on VGGSound-C. Detailed comparative results for this setting are provided in~\cref{exp:ks_v} and~\cref{exp:vgg_ks_a}.

\smallskip
\noindent \textbf{Continual Adaptation under Uni-Modal Corruption.} 
To evaluate continual adaptation within a single modality, we sequentially apply various types of corruption to one modality while keeping the others unaltered. The model must continuously adapt to these new corruptions without resetting parameters, demonstrating its ability to maintain robustness within the same modality. The comparison results for uni-modal continual corruption tasks are presented in~\cref{exp:ks_v_continual} and~\cref{exp:vgg_ks_a_continual}. Our method consistently outperforms others in the later stages, indicating superior robustness against catastrophic forgetting during continual adaptation.

\smallskip
\noindent \textbf{Continual Adaptation under Interleaved Modality Corruption.}
In this challenging scenario, corruption continuously alternates between different modalities. The model must adapt to these dynamically changing degradation sources across modalities while avoiding interference among them. Our proposed method demonstrates significantly improved performance under these interleaved modality distribution shifts (see~\cref{exp:ks_interleaved} and~\cref{exp:vgg_interleaved}).
This improvement results from our approach of externalizing domain-specific parameters, which are selectively activated and updated only upon detection of a distribution shift. This strategy ensures that source domain knowledge is preserved and not overwritten by domain-specific adaptations. Empirical results validate the effectiveness of this approach, with our method achieving average performance gains of 4.4\% and 1.5\% on the two benchmarks.

\begin{figure}[t]
    \centering
    \includegraphics[width=0.95\linewidth]{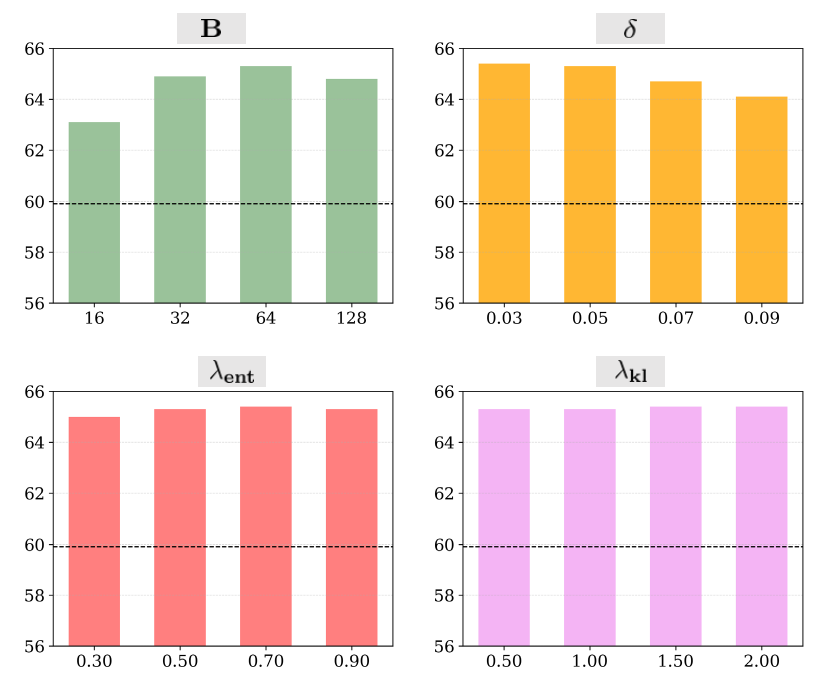}
    \caption{\textbf{Sensitivity Analysis of Hyper-parameters:} Batch Size ($\mathbf{B}$), Redundancy Threshold ($\mathbf{\delta}$) and Loss Coefficents ($\lambda_{\text{ent}}$, $\lambda_{\text{kl}}$).}
    \label{fig:sensitivity}
    \vspace{-6mm}
\end{figure}

\begin{figure}[t]
    \centering
    \includegraphics[width=0.95\linewidth]{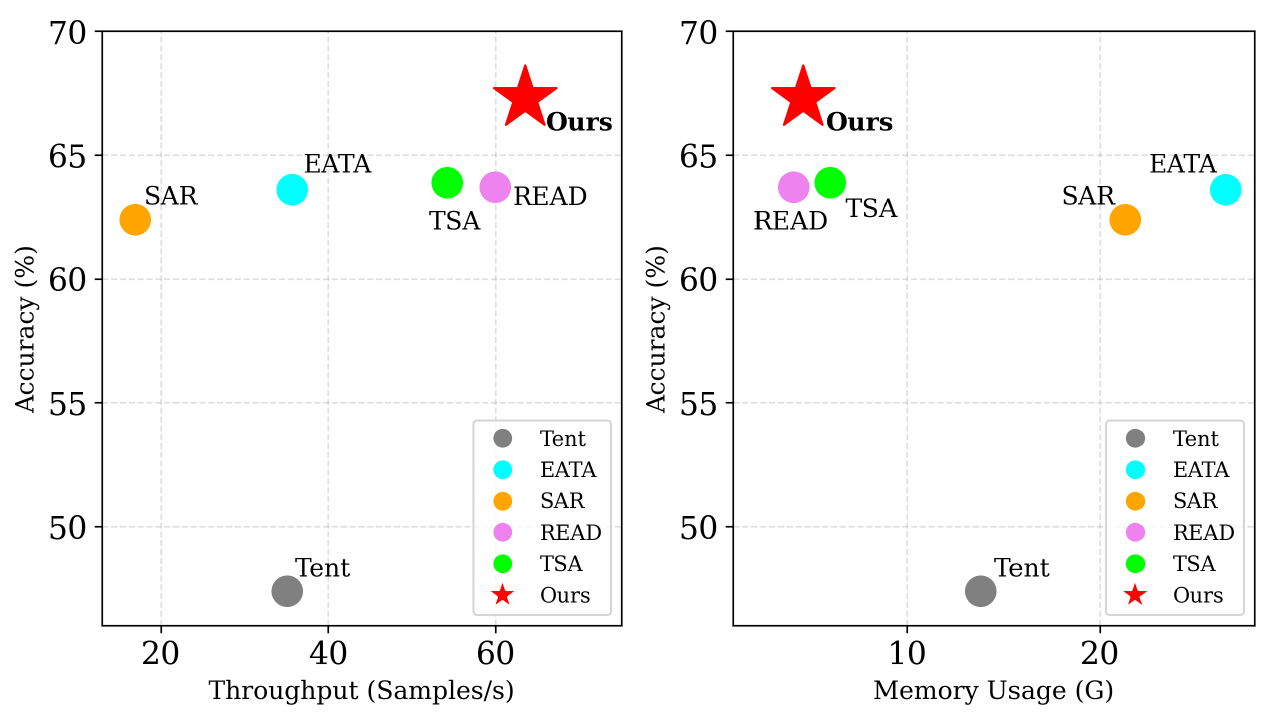}
    \caption{\textbf{Accuracy \vs Throughput and Memory Usage.} Compared to baselines, our method demonstrates superior performance with higher efficiency (observing comparable or lower computational cost and higher inference speed) on Kinetics50-C.}
    \label{fig:efficiency}
    \vspace{-6mm}
\end{figure}

\subsection{More Analysis}

\noindent \textbf{Component Analysis.} 
We conducted an ablation study (\cref{exp:ablation}) to evaluate the contribution of each component in our approach. Four experiments were designed: (i) without the stable adapter, (ii) without the plastic adapter, (iii) without asymmetric adaptation, and (iv) with asymmetric adaptation configured in the opposite manner. Removing either adapter resulted in decreased adaptive performance, indicating that the stable adapter is essential for extracting domain-invariant features and improving discrimination, which helps prevent negative transfer on the unbiased modality observed in other methods. Meanwhile, the plastic adapter provides necessary plasticity and domain-specific knowledge for effective target domain adaptation. Lastly, we assessed the asymmetric adaptation strategy by removing it, which caused a 5.9\% performance decline, and by applying the opposite configuration, which led to a 6.5\% decline. The substantial performance drops in both cases clearly demonstrate the effectiveness of our proposed asymmetric adaptation strategy.

\smallskip
\noindent \textbf{Sensitivity Analysis.} We performed a sensitivity analysis (\cref{fig:sensitivity}) on several key hyperparameters: batch size $B$, redundancy threshold $\delta$, and loss coefficients $\lambda_{\text{ent}}$ and $\lambda_{\text{kl}}$. Our method exhibits increased stability with larger batch sizes (32, 64, and 128), attributed to the integration of entropy minimization and diversity regularization. Additionally, the approach demonstrates low sensitivity to both the redundancy threshold and loss coefficients, significantly improving its practical applicability.

\smallskip
\noindent \textbf{Efficiency Analysis.} To confirm that our method does not introduce significant inference overhead during adaptation, we performed an efficiency analysis, measuring the number of parameters, memory usage, and inference speed (the number of samples processed per second). As shown in~\cref{fig:efficiency}, our approach successfully outperforms existing TTA methods in terms of accuracy while simultaneously maintaining higher operational efficiency.

\section{Conclusion}
We proposed DASP, a novel diagnose-then-mitigate framework that addresses the negative transfer and forgetting in MM-TTA. We base our approach on a key insight: a discrepancy in interdimensional feature redundancy can identify modality bias. DASP uses this insight by applying an asymmetric adaptation strategy with decoupled modality-specific adapters. This design preserves generalizable knowledge while allowing adaptation to distribution shifts in the biased modality. Our work underscores the limitations of modality-agnostic methods and shows that asymmetric strategy is a promising avenue for robust MM-TTA.

{
    \small
    \bibliographystyle{ieeenat_fullname}
    \bibliography{main}
}

\maketitlesupplementary

\noindent This appendix contains supplementary results and detailed analyses that further validate our approach. The material is organized as follows:
\begin{itemize}
    \item \cref{sec:exp_detail} offers detailed information regarding the benchmark and implementation.
    \item \cref{sec:ana_redu} provides additional evidence supporting our proposed redundancy score.
    \item  \cref{sec:suppl_exp} presents extended comparative experiments with recent MM-TTA methods.
\end{itemize}

\section{More Experimental Details}
\label{sec:exp_detail}

\subsection{Benchmarks}
\begin{figure}[t]
    \centering
    \includegraphics[width=\linewidth]{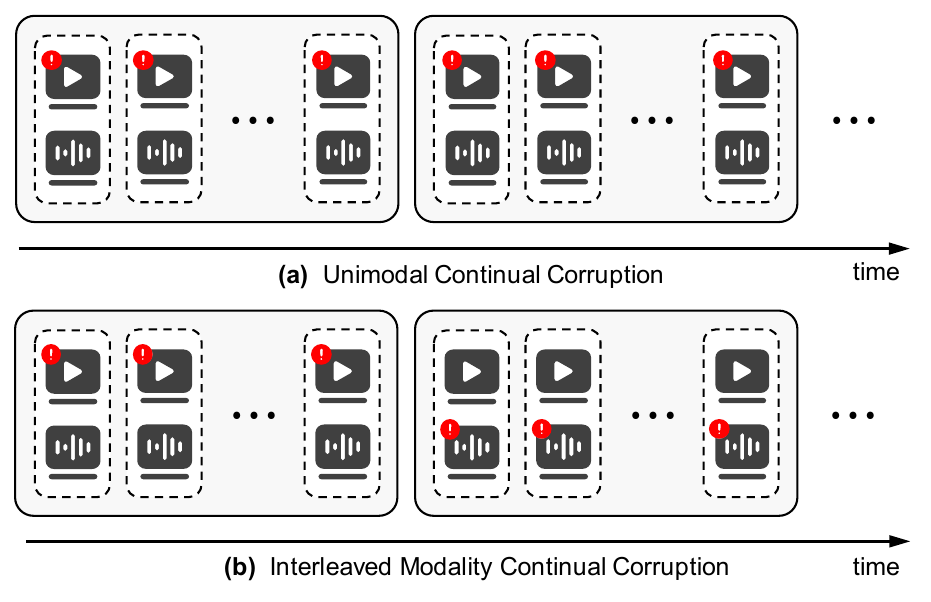}
    \caption{The illustration of uni-modal continual corruption and interleaved modality continual corruption.}
    \label{fig:interleaved}
\end{figure}

We construct two benchmarks based on Kinetics~\citep{kinetic} and VGGSound~\citep{vggsound}, to evaluate the performance of state-of-the-art methods under multi-modal domain shifts during test-time adaptation. We introduce three experimental setups: uni-modal episodic corruption, uni-modal continual corruption, and interleaved modality continual corruption. As illustrated in~\cref{fig:interleaved}, in the interleaved modality setup, corruption alternates continuously between different modalities (\eg from video to audio).

\noindent \textbf{Kinetics.} The Kinetics dataset is a large, high-quality benchmark used to recognize human actions in videos. It includes about 500,000 video clips covering 600 different action classes, with each class having at least 600 clips. Each clip is approximately 10 seconds long and labeled with a single action. The videos were collected from YouTube. Our study focuses on a subset of the Kinetics dataset, which contains 50 action classes and 2,466 test pairs.

\noindent \textbf{VGGSound.} The VGGSound dataset is a large-scale benchmark for audio-visual correspondence. It contains short audio clips extracted from YouTube videos recorded ``in the wild''. This ensures a clear match between the audio and visual content, with the sound source being visually identifiable. Each video in the dataset is 10 seconds long. We have collected 14,046 visual-audio pairs for testing.

\noindent \textbf{Kinetics50-C and VGGSound-C.} Following privious work~\citep{read}, we introduce 15 types of corruptions to the video modality data, including ``Gaussian Noise'', ``Shot Noise'', ``Impulse Noise'', ``Defocus Blur'', ``Glass Blur'', ``Motion Blur'', ``Zoom Blur'', ``Snow'', ``Frost'', ``Fog'', ``Brightness'', ``Contrast'', ``Elastic Transform'', ``Pixelate'', and ``JPEG Compression''. For the audio modality data, we introduce 6 types of corruptions, comprising ``Gaussian Noise'', ``Paris Traffic Noise'', ``Crowd Noise'', ``Rainy Noise'', ``Thunder Noise'' and ``Windy Noise''. Each corruption type is applied at five levels of severity.

\subsection{Implementations}
All baseline methods, along with our proposed approach, are optimized using the Adam optimizer, employing a learning rate of $1 \times 10^{-4}$ for both Kinetics-C and VGGSound-C.

\smallskip
\noindent \textbf{Tent.} For Tent~\citep{tent}, following its official implementation\footnote{\url{https://github.com/DequanWang/tent}}, we set the tunable parameters to those within the LayerNorm modules. \\
\textbf{EATA.} For EATA~\citep{eata}, following its official implementation\footnote{\url{https://github.com/mr-eggplant/EATA}}, we set the tunable parameters to those within the LayerNorm modules. Moreover, the exponential moving average (EMA) factor, cosine similarity threshold, and entropy threshold are set to 0.9, 0.05, and $0.4 \ln(C)$, respectively, where $C$ denotes the number of classes. \\ 
\textbf{READ.} For READ~\citep{read}, following its official implementation\footnote{\url{https://github.com/XLearning-SCU/2024-ICLR-READ}}, the tunable parameters are specified as the query, key, and value transformation matrices of the attention layer within the fusion block. \\
\textbf{TSA.} For TSA~\citep{tsa}, following its official implementation\footnote{\url{https://github.com/chenmc1996/Uni-Modal-Distribution-Shift}}, the modality-specific adapters are configured as the only tunable parameters. \\
\textbf{DASP (Ours).} Our proposed DASP adapts by updating two specialized modules: the stable and plastic adapters. The stable adapter uses a low-rank bottleneck design (rank $r=32$), defined as $h = x + W_{up}\sigma(W_{down}x)$, which acts as a structural regularizer to prevent overfitting. In contrast, the plastic adapter is a full-rank linear module (rank $r=768$), defined as $h = x + Wx$, providing enough capacity for feature realignment. Furthermore, we apply variance filtering by masking dimensions with near-zero variance ($\sigma^2 \approx 0$) before computing the redundancy score $R(\cdot)$, ensuring that invalid dimensions do not dilute the final score.

\section{Further Analysis of the Redundancy Score}
\label{sec:ana_redu}

\textbf{Theoretical Analysis.} The distribution shift is modeled as a low-rank perturbation in the latent space. For a perturbed sample $\tilde{\mathbf{z}} \in \mathbb{R}^D$, we consider the dominant rank-1 component: $\tilde{\mathbf{z}} = \mathbf{z} + \alpha\mathbf{v}$. To formalize our analysis, we establish the following \textbf{Assumptions}: \\
\textbf{1.} The dimensions of $\mathbf{z}$ are centered and uncorrelated, \ie, $\mathbb{E}[\mathbf{z}] = 0$ and $\mathrm{Cov}(\mathbf{z}) = \mathbf{I}_D$. \\
\textbf{2.} The shift intensity $\alpha$ is a random variable independent of $\mathbf{z}$, satisfying $\mathbb{E}[\alpha] = 0$ and $\mathrm{Var}(\alpha) = \sigma_\alpha^2 > 0$. \\
\textbf{3.} The shift direction $\mathbf{v}$ is non-sparse, such that $\|\mathbf{v}\|_0 \geq 2$. \\
\textbf{Theorem.} 
Let interdimensional redundancy be defined as $R(\mathbf{Z}) = \kappa \sum_{i \neq j} \mathcal{C}_{ij}^2$, where $\kappa = \frac{1}{D(D-1)} > 0$. Under Assumptions 1-3, the redundancy strictly increases under distribution shift, \ie, $R(\tilde{\mathbf{Z}}) > R(\mathbf{Z}) = 0$. \\
\textit{Proof.}
By the independence in Assumption 2, the covariance matrix of perturbed features $\tilde{\mathbf{Z}}$ is given by:
\begin{equation}
    \tilde{\Sigma} = \mathbb{E}[\tilde{\mathbf{z}}\tilde{\mathbf{z}}^\top] - \mathbb{E}[\tilde{\mathbf{z}}]\mathbb{E}[\tilde{\mathbf{z}}]^\top = \Sigma + \sigma_\alpha^2 \mathbf{v}\mathbf{v}^\top.
\end{equation}
From Assumption 1, we have $\Sigma = \mathbf{I}_D$. Thus, for any $i \neq j$, the $(i,j)$-th entry of $\tilde{\Sigma}$ is:
\begin{equation}
    \tilde{\Sigma}_{ij} = \Sigma_{ij} + \sigma_\alpha^2 v_i v_j = \sigma_\alpha^2 v_i v_j.
\end{equation}
The corresponding correlation coefficient $\tilde{\mathcal{C}}_{ij}$ is:
\begin{equation}
\tilde{\mathcal{C}}_{ij} = \frac{\sigma_\alpha^2 v_i v_j}{\sqrt{(1 + \sigma_\alpha^2 v_i^2)(1 + \sigma_\alpha^2 v_j^2)}}.
\end{equation}
According to Assumption 3, there exists at least one pair of indices $(m, n)$ with $m \neq n$ such that $v_m \neq 0$ and $v_n \neq 0$. Given $\sigma_\alpha^2 > 0$, it follows that $\tilde{\mathcal{C}}_{mn}^2 > 0$. We conclude:
\begin{equation}
    R(\tilde{\mathbf{Z}}) = \kappa \sum_{i \neq j} \tilde{\mathcal{C}}_{ij}^2 \geq \kappa (\tilde{\mathcal{C}}_{mn}^2 + \tilde{\mathcal{C}}_{nm}^2) > 0 = R(\mathbf{Z}).
\end{equation}
The proof is complete. \hfill $\square$

\begin{figure}[t]
    \centering
    \includegraphics[width=\linewidth]{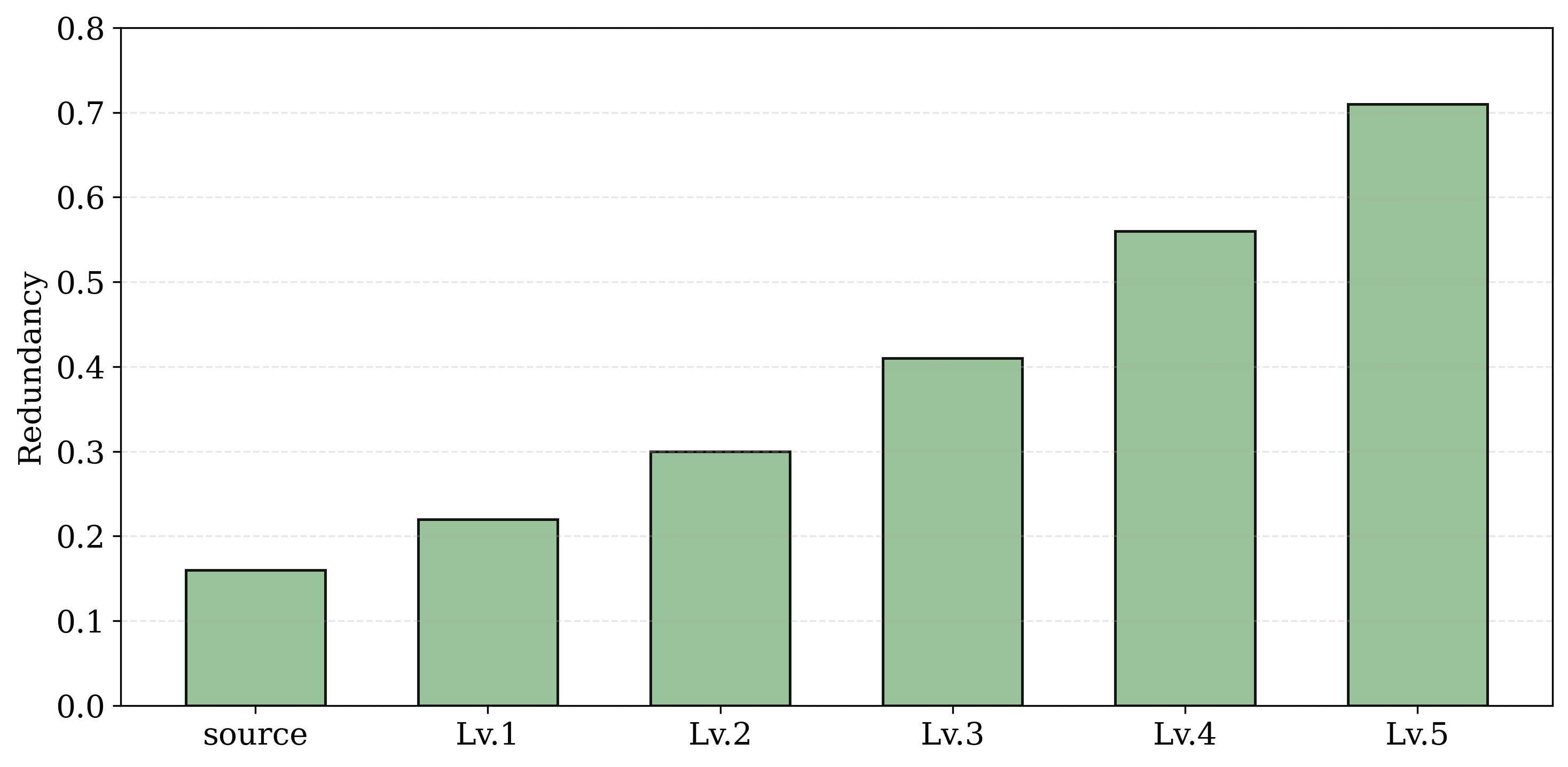}
    \caption{\textbf{Redundancy \vs Severity Level.} The redundancy score is evaluated across various severity levels of Kinetics50-C (Gaussian noise) and the original Kinetics50 validation set (source).}
    \label{fig:redu_vs_level}
\end{figure}

\begin{figure}[t]
    \centering
    \includegraphics[width=\linewidth]{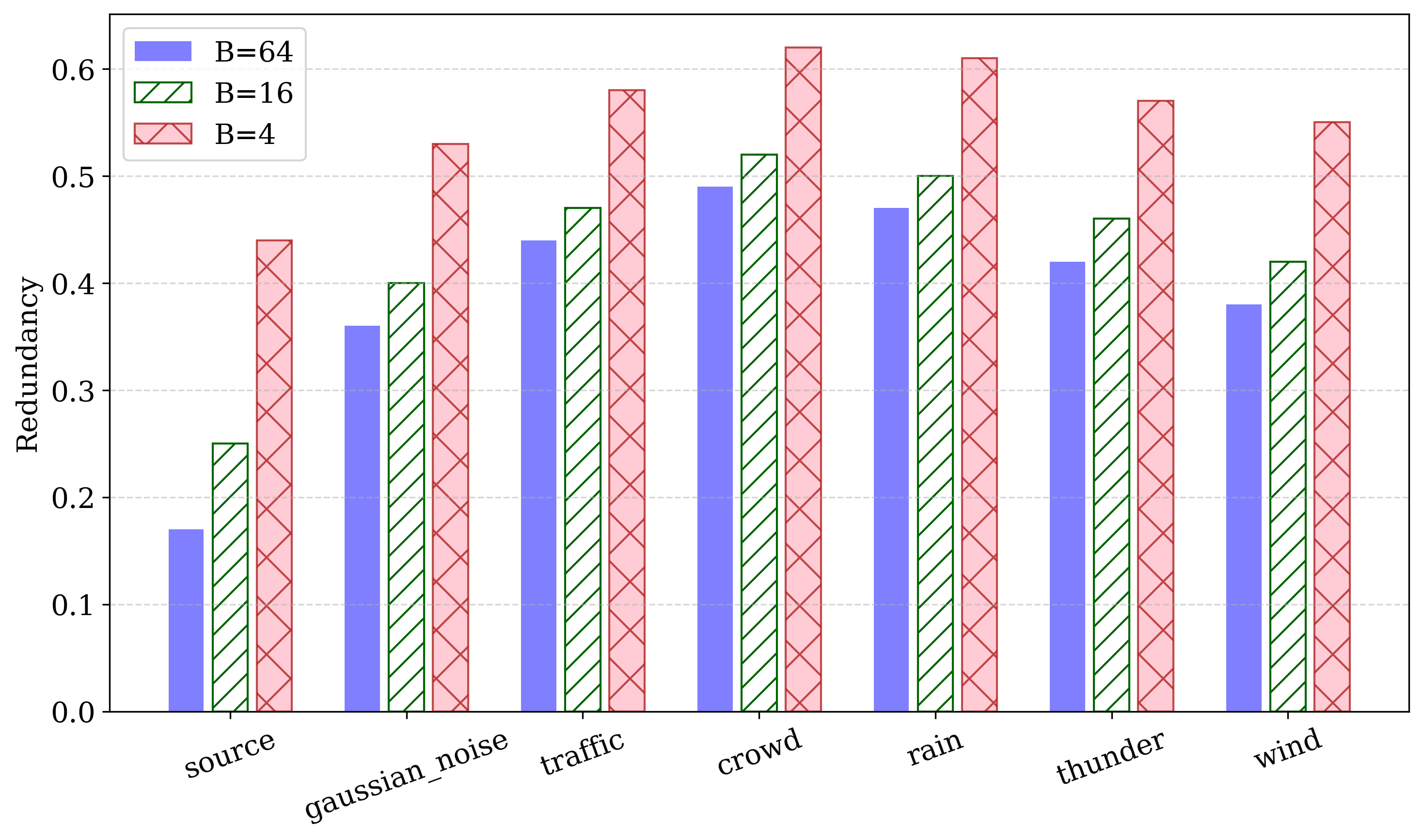}
    \caption{\textbf{Redundancy \vs Batch Size.} We investigate the correlation  between redundancy score and batch size on VGGSound-C with audio corruptions.}
    \label{fig:redu_vs_bs}
    \vspace{-5mm}
\end{figure}

\smallskip
\noindent \textbf{Empirical Analysis.} We present empirical results that validate the theoretical properties of $R(\mathbf{Z})$.

\smallskip
\noindent \textit{(i) Correlation with severity level.}
Our theoretical derivation establishes that the cross-dimensional correlation $\tilde{\mathcal{C}}_{ij}$ is fundamentally driven by the shift intensity variance $\sigma_\alpha^2$. In practical terms, an increase in $\sigma_\alpha^2$ directly corresponds to a higher severity level of the out-of-distribution (OOD) corruption. We empirically validate this direct relationship in \cref{fig:redu_vs_level}. The results demonstrate a clear, positive correlation between increasing corruption severity and the redundancy score $R(\mathbf{Z})$. This confirms our theoretical hypothesis: as inputs deviate further from the source manifold (higher $\sigma_\alpha^2$), the representation degradation exacerbates, which is precisely captured by the escalating redundancy score.

\smallskip
\noindent \textit{(ii) Correlation with batch size.}
The theoretical proof relies on the population covariance matrix $\tilde{\Sigma}$ calculated over the entire distribution via expectations ($\mathbb{E}$). However, in practice, $R(\mathbf{Z})$ acts as a statistical estimator computed over a finite batch of size $B$. \cref{fig:redu_vs_bs} assesses the stability of this sample estimator. While the absolute value of the score exhibits expected statistical variance when $B$ is small, the results validate a critical theoretical boundary: the redundancy under distribution shift $R(\tilde{\mathbf{Z}})$ remains consistently and strictly greater than that of the source domain $R(\mathbf{Z})$, echoing the $R(\tilde{\mathbf{Z}}) > R(\mathbf{Z})$ conclusion from our theorem. To further stabilize the estimate in scenarios with small $B$, we suggest caching samples (\eg via a momentum-based queue) to compute $R(\mathbf{Z})$ over a larger effective batch.

\begin{table*}[t]
    \caption{\textbf{Episodic Adaptation.} Comparison with SOTA methods on VGGSound-C with video corruptions (severity level 5) regarding \textbf{Accuracy (\%, $\uparrow$)}.}
    \label{exp:vgg_v}
    \centering
    \begin{threeparttable}
    \LARGE
    \renewcommand\arraystretch{1.2}
    \resizebox{\textwidth}{!}{
    \begin{tabular}{lcccccccccccccccc}
        \toprule
        \multicolumn{1}{c}{} & \multicolumn{3}{c}{Noise} & \multicolumn{4}{c}{Blur} & \multicolumn{4}{c}{Weather} & \multicolumn{4}{c}{Digital} & \multicolumn{1}{c}{} \\ 
        \cmidrule(lr){2-4} \cmidrule(lr){5-8} \cmidrule(lr){9-12} \cmidrule(lr){13-16}
        \multicolumn{1}{c}{Method} & Gauss. & Shot & Impul. & Defoc. & Glass & Mot. & Zoom & Snow & Frost & Fog & Brit. & Contr. & Elas. & Pix. & JPEG & \cellcolor{green!10} Avg. \\ 
        \midrule
        Source & 52.8 & 52.7 & 52.7 & 57.2 & 57.2 & 58.7 & 57.0 & 56.4 & 56.6 & 55.6 & 58.0 & 53.7 & 56.9 & 55.8 & 56.9 & \cellcolor{green!10} 56.0 \\
        $\bullet$ Tent \texttt{(ICLR'21)} & 52.7 & 52.7 & 52.7 & 56.7 & 56.7 & 57.9 & 57.1 & 55.9 & 56.3 & 56.3 & 58.4 & 54.0 & 57.4 & 56.2 & 56.7 & \cellcolor{green!10} 55.8 \\
        $\bullet$ EATA \texttt{(ICML'22)} & 53.0 & 52.8 & 53.0 & 57.0 & 57.0 & 58.1 & 57.2 & 56.3 & 56.8 & 56.8 & 58.7 & 54.1 & 57.6 & 56.4 & 57.0 & \cellcolor{green!10} 56.1 \\
        $\bullet$ SAR \texttt{(ICLR'23)} & 52.9 & 52.8 & 52.9 & 57.1 & 57.1 & 57.7 & 57.6 & 56.6 & 55.7 & 56.7 & 58.6 & 54.0 & 57.1 & 56.3 & 56.9 & \cellcolor{green!10} 56.0 \\
        $\bullet$ READ \texttt{(ICLR'24)} & 53.6 & 53.6 & 53.5 & 57.9 & 57.7 & 59.4 & 58.8 & 56.8 & 57.1 & 56.9 & 59.9 & 53.8 & 58.6 & 57.1 & 57.6 & \cellcolor{green!10} 56.9 \\
        $\bullet$ TSA \texttt{(ICML'25)} & 53.4 & 53.4 & 53.1 & 57.5 & 57.3 & 58.8 & 58.3 & 56.6 & 56.9 & 57.0 & 59.3 & 55.3 & 58.0 & 56.7 & 57.8 & \cellcolor{green!10} 56.6 \\
        \rowcolor{blue!10}  
        \textbf{$\bullet$ Ours} & \textbf{54.7} & \textbf{54.7} & \textbf{54.6} & \textbf{58.3} & \textbf{58.3} & \textbf{59.5} & \textbf{59.0} & \textbf{57.3} & \textbf{58.0} & \textbf{57.8} & \textbf{60.1} & \textbf{56.0} & \textbf{58.9} & \textbf{57.4} & \textbf{58.0} & \textbf{57.5} \\ 
        \bottomrule
    \end{tabular}
    }
    \end{threeparttable}
\end{table*}

\begin{table*}[t]
    \caption{\textbf{Continual Adaptation.} Comparison with SOTA methods on VGGSound-C with video corruptions (severity level 5) regarding \textbf{Accuracy (\%, $\uparrow$)}.}
    \label{exp:vgg_v_continual}
    \centering
    \begin{threeparttable}
    \LARGE
    \renewcommand\arraystretch{1.2}
    \resizebox{\textwidth}{!}{
    \begin{tabular}{lcccccccccccccccc}
        \toprule
        \multicolumn{1}{c}{} & \multicolumn{3}{c}{Noise} & \multicolumn{4}{c}{Blur} & \multicolumn{4}{c}{Weather} & \multicolumn{4}{c}{Digital} & \multicolumn{1}{c}{} \\ 
        \cmidrule(lr){2-4} \cmidrule(lr){5-8} \cmidrule(lr){9-12} \cmidrule(lr){13-16}
        \multicolumn{1}{c}{Method} & Gauss. & Shot & Impul. & Defoc. & Glass & Mot. & Zoom & Snow & Frost & Fog & Brit. & Contr. & Elas. & Pix. & JPEG & \cellcolor{green!10} Avg. \\ 
        \midrule
        \multicolumn{1}{c}{} &
        \multicolumn{15}{c}{$t \; \xrightarrow{\hspace{23cm}}$} \\ 
        \midrule
        Source & 52.8 & 52.7 & 52.7 & 57.2 & 57.2 & 58.7 & 57.0 & 56.4 & 56.6 & 55.6 & 58.0 & 53.7 & 56.9 & 55.8 & 56.9 & \cellcolor{green!10} 56.0 \\
        $\bullet$ Tent \texttt{(ICLR'21)} & 52.7 & 52.4 & 51.5 & 54.4 & 54.0 & 55.2 & 54.8 & 52.1 & 52.9 & 52.8 & 52.8 & 50.6 & 52.7 & 52.2 & 51.8 & \cellcolor{green!10} 52.9 \\
        $\bullet$ EATA \texttt{(ICML'22)} & 53.0 & 53.4 & 53.5 & 56.7 & 57.1 & 58.6 & 58.5 & 55.8 & 56.8 & 57.3 & 58.0 & 55.3 & 57.6 & 57.1 & 57.5 & \cellcolor{green!10} 56.4 \\
        $\bullet$ SAR \texttt{(ICLR'23)} & 52.9 & 53.0 & 53.0 & 56.9 & 56.7 & 58.5 & 57.5 & 56.0 & 56.8 & 56.0 & 58.2 & 54.1 & 57.1 & 55.4 & 56.3 & \cellcolor{green!10} 55.9 \\
        $\bullet$ READ \texttt{(ICLR'24)} & 53.7 & 54.0 & 53.9 & 57.4 & 57.4 & 58.2 & 57.9 & 56.5 & 57.1 & 57.0 & 58.0 & 55.4 & 56.8 & 56.2 & 56.2 & \cellcolor{green!10} 56.4 \\
        $\bullet$ TSA \texttt{(ICML'25)} & 53.4 & 53.8 & 53.7 & 57.1 & 57.4 & 58.5 & 58.1 & 56.5 & 57.7 & 57.2 & 58.7 & 55.5 & 57.6 & 56.5 & 57.1 & \cellcolor{green!10} 56.6 \\
        \rowcolor{blue!10} 
        \textbf{$\bullet$ Ours} & \textbf{54.0} & \textbf{54.9} & \textbf{54.9} & \textbf{58.3} & \textbf{58.8} & \textbf{59.9} & \textbf{59.5} & \textbf{57.5} & \textbf{58.2} & \textbf{58.3} & \textbf{60.0} & \textbf{56.4} & \textbf{58.6} & \textbf{57.7} & \textbf{58.3} & \textbf{57.7} \\ 
        \bottomrule
    \end{tabular}
    }
    \end{threeparttable}
\end{table*}

\begin{table*}[!h]
    \caption{\textbf{Episodic Adaptation.} Comparison with MM-TTA methods on Kinetics50-C with video corruptions (severity level 5) regarding \textbf{Accuracy (\%, $\uparrow$)}.}
    \label{exp:mmtta}
    \centering
    \begin{threeparttable}
    \LARGE
    \renewcommand\arraystretch{1.2}
    \resizebox{\textwidth}{!}{
    \begin{tabular}{lcccccccccccccccc}
        \toprule
        \multicolumn{1}{c}{} & \multicolumn{3}{c}{Noise} & \multicolumn{4}{c}{Blur} & \multicolumn{4}{c}{Weather} & \multicolumn{4}{c}{Digital} & \multicolumn{1}{c}{} \\ 
        \cmidrule(lr){2-4} \cmidrule(lr){5-8} \cmidrule(lr){9-12} \cmidrule(lr){13-16}
        \multicolumn{1}{c}{Method} & Gauss. & Shot & Impul. & Defoc. & Glass & Mot. & Zoom & Snow & Frost & Fog & Brit. & Contr. & Elas. & Pix. & JPEG & \cellcolor{green!10} Avg. \\ 
        \midrule
        Source & 46.8 & 48.0 & 46.9 & 67.5 & 62.2 & 70.8 & 66.7 & 61.6 & 60.3 & 46.7 & 75.2 & 52.1 & 65.7 & 66.5 & 61.9 & \cellcolor{green!10} 59.9 \\
        $\bullet$ READ \texttt{(ICLR'24)} & 49.4 & 49.7 & 49.0 & 68.0 & 65.1 & 71.2 & 69.0 & 64.5 & 64.4 & 57.4 & 75.5 & 53.6 & 68.3 & 68.0 & 65.1 & \cellcolor{green!10} 62.5 \\
        $\bullet$ ABPEM \texttt{(AAAI'25)} & 50.6 & 51.1 & 50.5 & 68.7 & 66.6 & \textbf{72.6} & 69.6 & 64.4 & 66.2 & 60.5 & 76.0 & 55.3 & 69.5 & 69.2 & 66.2 & \cellcolor{green!10} 63.8 \\
        $\bullet$ SuMi \texttt{(ICLR'25)} & 50.1 & 50.7 & 50.4 & 68.2 & 65.6 & 72.2 & 69.7 & 65.7 & 67.0 & 56.5 & \textbf{77.1} & 55.2 & 69.3 & 71.2 & \textbf{68.9} & \cellcolor{green!10} 63.9 \\
        $\bullet$ TSA \texttt{(ICML'25)} & 50.7 & 51.1 & 50.4 & 67.9 & 67.1 & 71.7 & 69.2 & 65.5 & 66.2 & 61.3 & 75.2 & 56.2 & 69.5 & 68.8 & 66.6 & \cellcolor{green!10} 63.8 \\
        $\bullet$ BriMPR \texttt{(AAAI'26)} & 50.0 & 50.8 & 50.3 & 68.4 & 67.5 & 71.4 & 69.0 & 65.3 & 65.1 & 63.4 & 76.1 & 56.8 & \textbf{71.6} & \textbf{72.2} & 67.2 & \cellcolor{green!10} 64.3 \\
        \rowcolor{blue!10} 
        \textbf{$\bullet$ Ours} & \textbf{50.8} & \textbf{51.6} & \textbf{50.7} & \textbf{70.2} & \textbf{69.3} & 72.3 & \textbf{71.3} & \textbf{66.1} & \textbf{68.2} & \textbf{63.5} & 75.2 & \textbf{58.1} & 71.2 & 70.5 & 68.6 & \textbf{65.2} \\ 
        \bottomrule
    \end{tabular}
    }
    \end{threeparttable}
\end{table*}

\section{Extended Comparative Experiments}
\label{sec:suppl_exp}
\textbf{Main experiments.}
We report additional results for the main experiments that were not included in the main text. These results cover the evaluation of episodic and continual adaptation on VGGSound-C with video corruptions. As the video serves as an auxiliary modality in the VGGSound dataset, the improvements observed across all methods are modest. Nonetheless, as demonstrated in~\cref{exp:vgg_v,exp:vgg_v_continual}, our approach surpasses the current state-of-the-art methods by 0.6\% and 1.1\%, respectively.

\smallskip
\noindent \textbf{Comparison with MM-TTA baselines.}
In the comparative experiments presented in the main text, we focus only on the READ~\citep{read} and TSA~\cite{tsa} methods within MM-TTA. To further demonstrate the superiority of our approach, we have included additional recently proposed baselines: \ding{182} \textbf{ABPEM}~\citep{abpem}, which introduces attention bootstrapping and master entropy minimization to reduce the attention gap. \ding{183} \textbf{SuMi}~\citep{sumi} which proposes two novel strategies: sample identification with interquartile range smoothing and unimodal assistance, and mutual information sharing. \ding{184} \textbf{BriMPR}~\citep{brimpr}, which employs a progressive re-alignment method to address the coupling between uni-modal and multi-modal misalignments. The experiment is performed under consistent conditions, specifically a learning rate of 1e-4 and a batch size of 64, to ensure a fair comparison. The implementation follows their official codes~\footnote{\url{https://github.com/YushengZhao/ABPEM}}~\footnote{\url{https://github.com/zrguo/SuMi}}~\footnote{\url{https://github.com/Luchicken/BriMPR}}. It should be noted that the BriMPR method updates the modality-specific encoders through prompt learning, which incurs significant computational costs, and also utilizes source domain statistics. However, as demonstrated in~\cref{exp:mmtta}, our method surpasses BriMPR, the second-best approach, by 0.9\%, highlighting the superior effectiveness.

\end{document}